\documentclass[lettersize,journal]{IEEEtran}
\usepackage{amsmath,amsfonts}
\usepackage{array}
\usepackage[caption=false,font=normalsize,labelfont=sf,textfont=sf]{subfig}
\usepackage{textcomp}
\usepackage{stfloats}
\usepackage{url}
\usepackage{verbatim}
\usepackage{graphicx}
\usepackage{booktabs}
\usepackage{multirow}
\usepackage[mathscr]{eucal}
\usepackage{cite}
\usepackage{tcolorbox}
\usepackage{hyperref}
\hypersetup{hidelinks,
	colorlinks=true,
	allcolors=black,
	pdfstartview=Fit,
	breaklinks=true}
\usepackage{bbding}
\usepackage{xcolor}
\usepackage{colortbl}
\definecolor{mypurple}{HTML}{E8EEDD}
\usepackage{pifont}

\usepackage{makecell}
\usepackage[ruled,vlined,linesnumbered,noresetcount]{algorithm2e}
\SetKwInput{kwInit}{Init}
\SetKwComment{Comment}{$\triangleright $}{}

\newcommand{\modelname}{FedDCL} 

\begin{document}


\title{Data-Free Continual Learning of Server Models in Model-Heterogeneous Cloud-Device Collaboration}



\author{Xiao Zhang, \and Zengzhe Chen, \and Yuan Yuan, \and Yifei Zou, \and Fuzhen Zhuang, \and Wenyu Jiao, \and Yuke Wang,  \and and Dongxiao Yu

\thanks{Xiao Zhang, Zengzhe Chen, Yifei Zou and Dongxiao Yu are with the School of Computer Science and Technology, Shandong University, Qingdao 266237, China. (Dongxiao Yu is the corresponding author.)}
\thanks{Yuan Yuan is with the School of Software $\&$ Joint SDU-NTU Centre for Artificial Intelligence Research (C-FAIR), Shandong University, Jinan, 250000, China.}
\thanks{Fuzhen Zhuang is with the Institute of Artificial Intelligence, SKLSDE,
School of Computer Science, Beihang University, Beijing 100191, China.}
\thanks{Wenyu Jiao and Yuke Wang are with the Desautels Faculty of Management, McGill University, Montréal, Canada.} 
}

\markboth{Journal of \LaTeX\ Class Files,~Vol.~14, No.~8, August~2021}%
{Shell \MakeLowercase{\textit{et al.}}: A Sample Article Using IEEEtran.cls for IEEE Journals}


\maketitle
\begin{abstract}
The rise of cloud-device collaborative computing has enabled intelligent services to be delivered across distributed edge devices while leveraging centralized cloud resources. In this paradigm, federated learning (FL) has become a key enabler for privacy-preserving model training without transferring raw data from edge devices to the cloud. However, with the continuous emergence of new data and increasing model diversity, traditional federated learning faces significant challenges, including inherent issues of data heterogeneity, model heterogeneity and catastrophic forgetting, along with new challenge of knowledge misalignment.
In this study, we introduce \modelname, a novel framework designed to enable data-free continual learning of the server model in a model-heterogeneous federated setting. We leverage pre-trained diffusion models to extract lightweight class-specific prototypes, which confer a threefold data-free advantage, enabling: (1) generation of synthetic data for the current task to augment training and counteract non-IID data distributions; (2) exemplar-free generative replay for retaining knowledge from previous tasks; and (3) data-free dynamic knowledge transfer from heterogeneous devices to the cloud server.
Experimental results on various datasets demonstrate the effectiveness of \modelname, showcasing its potential to enhance the generalizability and practical applicability of federated cloud-device collaboration in dynamic settings.
\end{abstract}


\begin{figure}[t]
    \centering
    \begin{minipage}[t]{1\linewidth}
    \centering
        \begin{tabular}{@{\extracolsep{\fill}}c@{}c@{}c@{}@{\extracolsep{\fill}}}
            \includegraphics[width=1\linewidth]{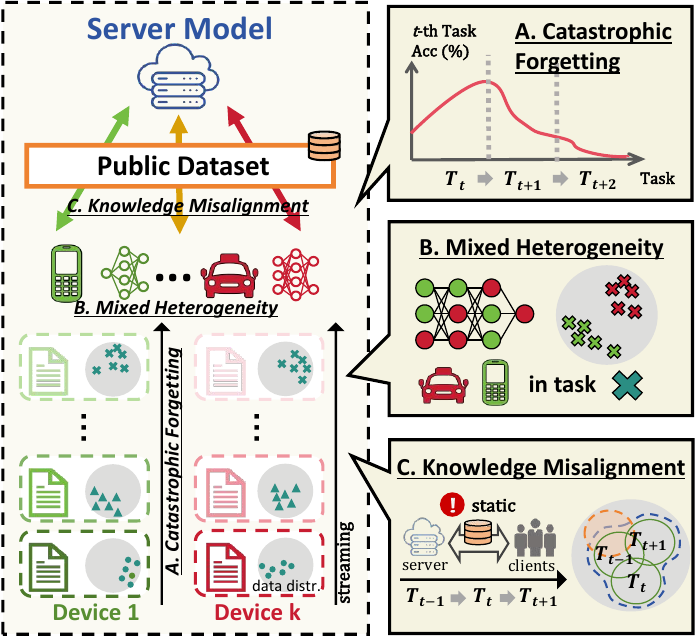}\\
        \end{tabular}
    \end{minipage}
    \caption{Illustration of the three main challenges in heterogeneous federated continual learning: (1) Catastrophic forgetting, where sequential task learning overwrites past knowledge and degrades earlier task performance; (2) Model heterogeneity, where diverse device architectures hinder direct parameter aggregation; and (3) Knowledge misalignment, where public dataset, due to static nature, fail to align with evolving task domains to transfer knowledge dynamically.
    }
    \label{fig:problem}
\end{figure}

\section{Introduction}

\IEEEPARstart{I}{n} recent years, the proliferation of edge devices, such as smartphones, IoT sensors, and surveillance cameras, has driven a paradigm shift toward cloud-device collaborative computing \cite{Cloud_edge_intro}. Within this ecosystem, Federated Learning (FL) \cite{Fedavg} has emerged as a representative machine learning paradigm that enables collaborative model training across multiple entities while preserving data privacy by keeping data localized. This design is strongly motivated by stringent data protection regulations, such as the GDPR \cite{gdpr} and CCPA \cite{ccpa}, which impose strict limitations on the centralized collection and processing of personal data.

However, in real-world applications where data rapidly evolve, the classic FL framework faces significant challenges, such as catastrophic forgetting\cite{catastrophic_forgetting}. 
For better collecting, accumulating and leveraging knowledge from streaming data, federated continual learning (FCL) has recently attracted growing attention \cite{FCL_survey1}\cite{FCL_survey2}. 
The FCL paradigm enables models to continuously adapt to new tasks or data distributions in the dynamic environment while retaining previously acquired knowledge.
This capability has led to successful applications in multiple fields, such as action recognition\cite{application_ar}, financial audit \cite{application_fin}, person re-identification\cite{application_id}, and network traffic classification\cite{trafic}\cite{trafic2}. In a typical FCL setting, the server generally assumes that all devices employ a completely homogeneous model architecture and communicates through parameter aggregation, as seen in methods like 
FedWeIT\cite{Fedwelt}, FedKNOW\cite{fedknow},
TARGET\cite{target},
and FedCIL\cite{fedcil}.
However, with the widespread adoption of mobile and IoT devices, models have become increasingly diverse in architecture and capacity. 
For example, some large pretrained models deployed on cloud servers, such as BERT\cite{bert}, ViT\cite{vit}, and even modern large language models\cite{touvron2023llama}\cite{achiam2023gpt}, often possess orders of magnitude more parameters than models on edge devices.
This growing disparity in model scale and structure introduces significant model heterogeneity, which hinders effective collaboration towards obtaining a high-quality server model and ultimately limits the applicability and generalizability of FCL in real-world, heterogeneous environments.

To bridge the barrier of architectural differences, knowledge distillation\cite{hinton2015distilling} enables efficient knowledge exchange by communicating soft predictions.
This approach has been adapted to the federated setting through federated distillation (FD), offering a compelling alternative to conventional parameter aggregation for knowledge exchange\cite{FD}.
Methods like FedGems\cite{fedgems}, Fed-ET\cite{Fedet}, FedMKT\cite{fan2024fedmkt}, and FedBiOT\cite{wu2024fedbiot} achieve heterogeneous model collaboration with an auxiliary dataset, bridging the model architecture gap. 
However, the above mentioned works are all limited to the single, static task. 
Although studies like FLwF-2T\cite{fedLWF2T} and CFeD\cite{ma2022continual} combine distillation with dynamic settings in FCL, they primarily treat distillation as a regularization technique to mitigate forgetting, completely overlooking its potential for heterogeneous model collaboration. 
Therefore, the scenarios of leveraging heterogeneous models deployed across diverse devices to collectively and continuously train a larger server model in streaming knowledge remain an under-explored area.

\begin{figure}[t]
    \centering
    \begin{minipage}[t]{1\linewidth}
    \centering
        \begin{tabular}{@{\extracolsep{\fill}}c@{}c@{}c@{}@{\extracolsep{\fill}}}
            \includegraphics[width=1\linewidth]{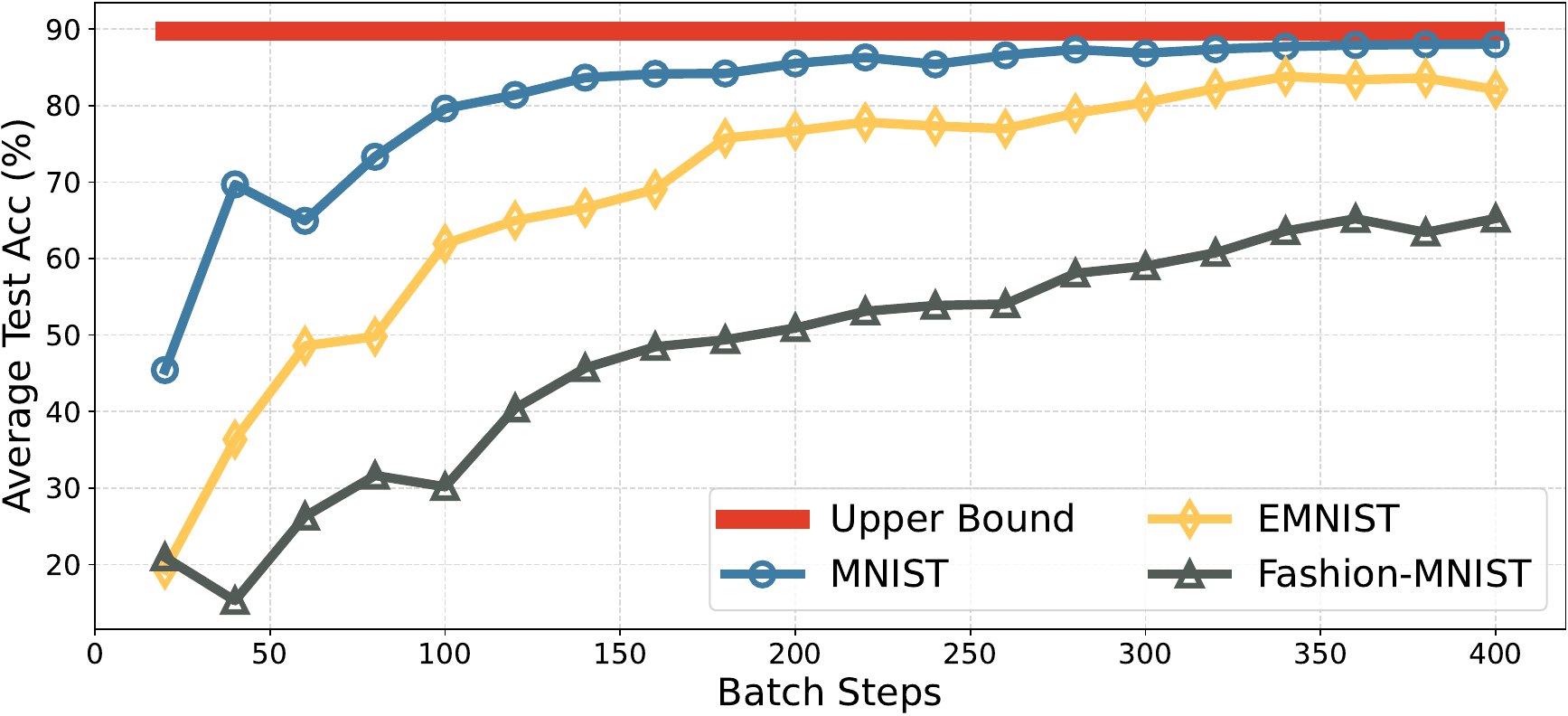}\\
        \end{tabular}
    \end{minipage}
    \caption{The performance of a student model distilling knowledge from a teacher on MNIST using different public datasets. Redline means the teacher's accuracy, which can be seen as upper bound.}
    \label{fig:data_align}
 \end{figure}

As shown in Figure.~\ref{fig:problem}, heterogeneous federated continual learning faces three main challenges. 
First, the problem stems from \textbf{catastrophic forgetting} of server models in continual learning. As clients sequentially learn from non-stationary, streaming data over time, previously acquired knowledge tends to be overwritten or degraded, leading to performance collapse on earlier tasks. 
Then, \textbf{model heterogeneity} presents another major obstacle. Due to resource constraints, participating devices may deploy vastly different model architectures (e.g., varying depths, widths, or even model designs), which complicates knowledge transfer. Conventional federated parameter aggregation methods become ineffective under such architectural diversity.
Finally, beyond the above inherent challenges posed by federated distillation and federated continual learning, their intersection introduces a novel challenge, which we term \textbf{knowledge misalignment}. The rapidly evolving data environment exacerbates the limitations of static public datasets commonly used in knowledge distillation. As illustrated in the toy example in Figure.~\ref{fig:data_align}, the effectiveness of distillation heavily depends on the domain alignment between the auxiliary dataset and the target task\cite{su2022domain}. Manually select high-quality auxiliary datasets may expose sensitive domain information and compromising privacy. Moreover, in scenarios where domain knowledge evolves rapidly, the static nature of such datasets often struggles to accurately capture the dynamic changes in domain knowledge, thus hindering effective knowledge transfer \cite{zhang2023towards}. 
Therefore, a practical problem arises: \textit{how can we enable heterogeneous device models to collaboratively and continuously learn a server model from streaming knowledge in a data-free manner?}

Along this line, we propose \modelname, a novel framework designed to facilitate knowledge collaboration among heterogeneous models to continuously train a server model under sequentially arriving tasks.
We achieve data-free dynamic knowledge alignment in both knowledge transfer and knowledge retention. 
Specifically, \modelname \ employs pre-trained diffusion models to generate compact, class-specific prototypes that capture the essential features of each class.
This design offers three-fold benefits: (1) We can augment data for the current task through prototypes to strengthen training and mitigate non-IID challenges. (2) By replaying previously learned knowledge through synthesized samples and combining them with incoming real data during training, \modelname \ effectively mitigates catastrophic forgetting and combats non-IID data shifts, all while avoiding the storage of any raw exemplars. (3) By adaptively generating synthetic data aligned with evolving task distributions, \modelname \ enables seamless collaboration among heterogeneous devices, eliminating dependence on static public dataset.
For the server model, we employ a multi-teacher distillation strategy, where knowledge from the latest task is transferred from edge clients and past models serve as teachers to mitigate catastrophic forgetting, effectively enabling the server model to continuously acquire knowledge.
We conduct extensive experiments with various settings on two datasets, revealing the superiority of \modelname \ over other baselines. Our code is available at: \url{https://anonymous.4open.science/r/FCL-E7B9}. Our main contributions can be summarized as follows:
\begin{itemize}
\item We propose a novel federated continual learning framework that enables heterogeneous devices to collaboratively train a central server model under streaming tasks, overcoming the hybrid heterogeneity of models and data, catastrophic forgetting, and knowledge misalignment caused by continuously evolving task knowledge.

\item We introduce lightweight class prototypes generated by a pre-trained diffusion model to enable data-free knowledge distillation and exemplar-free knowledge replay, removing the reliance on static auxiliary datasets while dynamically aligning with evolving task distributions.

\item We conduct extensive experiments with various settings across two datasets. Results reveal that \modelname \ achieves a better performance, improving the server accuracy by 9.00\%/6.48\%/23.67\%/16.40\% on four settings.

\end{itemize}

\section{Related Work}
\subsection{Model-Heterogeneous Federated Learning.}
Federated learning is designed for privacy-sensitive scenarios, enabling collaborative training among decentralized clients without centralizing raw data.
Classical federated learning algorithms, exemplified by FedAvg \cite{Fedavg} and most of its variants\cite{FedMoon}\cite{FedProx}, aggregate locally updated model parameters from participating devices to construct and refine a global server model.
Despite the notable success of homogeneous federated learning, traditional strategies that assume uniform model architectures are becoming increasingly impractical as architectural heterogeneity grows more prevalent\cite{heterogeneous_survey1}.
To tackle the bottleneck of training a global model with heterogeneous clients, existing model-heterogeneous federated learning approaches primarily follow two technical paradigms \cite{heterogeneous_survey2}: partial-training-based methods and knowledge-distillation-based methods. 
1) Partial-training-based methods refer to approaches in which the server model is compressed to yield lightweight variants for localized training. Common approaches include quantization \cite{ozkara2021quped}\cite{markov2023quantized}, which lowers numerical precision of weights, and pruning \cite{diao2020heterofl}\cite{jiang2022model}\cite{zhu2022resilient}\cite{wang2023theoretical}\cite{cho2024heterogeneous}, which removes redundant parameters or structures.
2) Knowledge-distillation-based methods typically transfer knowledge using soft predictions or feature representations\cite{hinton2015distilling}, without requiring architectural alignment.

\subsection{Federated Distillation.}
We focus on knowledge-distillation-based methods, which support higher degrees of model heterogeneity and are more common in practice. In this paradigm, clients train heterogeneous models locally and upload distilled knowledge. 
Methods like FedMD\cite{li2019fedmd}, DS-FL\cite{itahara2021distillation}, FedDF\cite{lin2020ensemble}, KT-pFL\cite{KT-pFL} and pFedHR\cite{pFedHR} enable communication among heterogeneous clients through a proxy dataset.
Additionally, as the model size on the server side continues to grow, some works like FedGems\cite{fedgems}, FedKEMF\cite{FedKEMF}, Fed-ET\cite{Fedet} and FedMKT\cite{fan2024fedmkt} focus on utilizing clients to collaboratively train a larger server model. 
Most existing methods overly rely on a fixed public dataset that is closely aligned with the target domain to facilitate effective knowledge exchange. However, obtaining such a dataset is often unavailable in practice due to strict data privacy constraints.
Although studies such as DaFKD \cite{wang2023dafkd} and PeFAD \cite{xu2024pefad} do not rely on public datasets but instead generate synthetic data resembling the target distribution to enhance knowledge distillation, these data-free distillation methods are primarily designed to assist model-homogeneous federated learning.
Besides, model-heterogeneous federated learning methods like FedGEN\cite{zhu2021data} FedGD\cite{zhang2023towards} and FedZKT\cite{fedzkt} train a global generator to produce synthetic data for replacing public data. However, training a generator from scratch is often time-consuming and the process may be unstable\cite{gan_collapse}.
Last but not least, the above generative-based methods do not account for the dynamic matching process with continuous knowledge, rendering them ineffective in our context.

\subsection{Federated Continual Learning.} 
Continual learning aims to train models sequentially on task flow while retaining previously acquired knowledge, a challenge commonly referred to as catastrophic forgetting\cite{cl_survey}.
Building upon this, Federated continual learning (FCL) has emerged as a promising paradigm that enables distributed clients to continuously collaborate while preserving privacy. 
Inspired by classic continual learning, FedWeIT\cite{Fedwelt} introduces the concept of federated continual learning. However, it focuses solely on the clients without considering the server model and necessitates the ID of the task for inference.
Regularization-based methods, such as FLwF-2T\cite{fedLWF2T} and CFeD\cite{ma2022continual} , mitigate catastrophic forgetting by penalizing parameter deviations through a knowledge distillation loss term. 
Data replay methods generally achieve stronger performance. Methods like FedKNOW\cite{fedknow} and FedViT\cite{fedvit} directly store real data from past tasks for replay, explicitly helping the model to recall previously learned knowledge. However, such direct storage of raw data may violate privacy constraints.
Inspired by DGR\cite{shin2017continual}, recent works achieve privacy-preserving data replay through synthetic data. 
Methods such as MFCL\cite{babakniya2024data}, TARGET\cite{target}, and FedCIL\cite{fedcil} employ adversarial training to learn generators that synthesize data for preserving previously acquired knowledge. 
Similarly, FedCBC\cite{yu2024overcoming} addresses spatial-temporal catastrophic forgetting via a VAE-based framework, yet it constructs separate binary classifiers for each class, limiting the flexibility. DDDR\cite{dddr} leverages pre-trained diffusion models to perform class inversion for knowledge retention.
It is worth noting that these methods are not designed to support collaboration among heterogeneous models. 
The recent work Mufti\cite{mufti} explores continual collaboration among heterogeneous clients through the use of a public dataset. However, introducing such a dataset raises privacy leakage concerns. Moreover, it does not consider the server side model.

Our approach significantly differs from the above methods, as it aims to simultaneously explore data-free continual learning while enabling data-free collaboration among heterogeneous models.

\begin{figure*}[t]
    \centering
    \begin{minipage}{1\linewidth}
    \centering
            \includegraphics[width=1\linewidth]{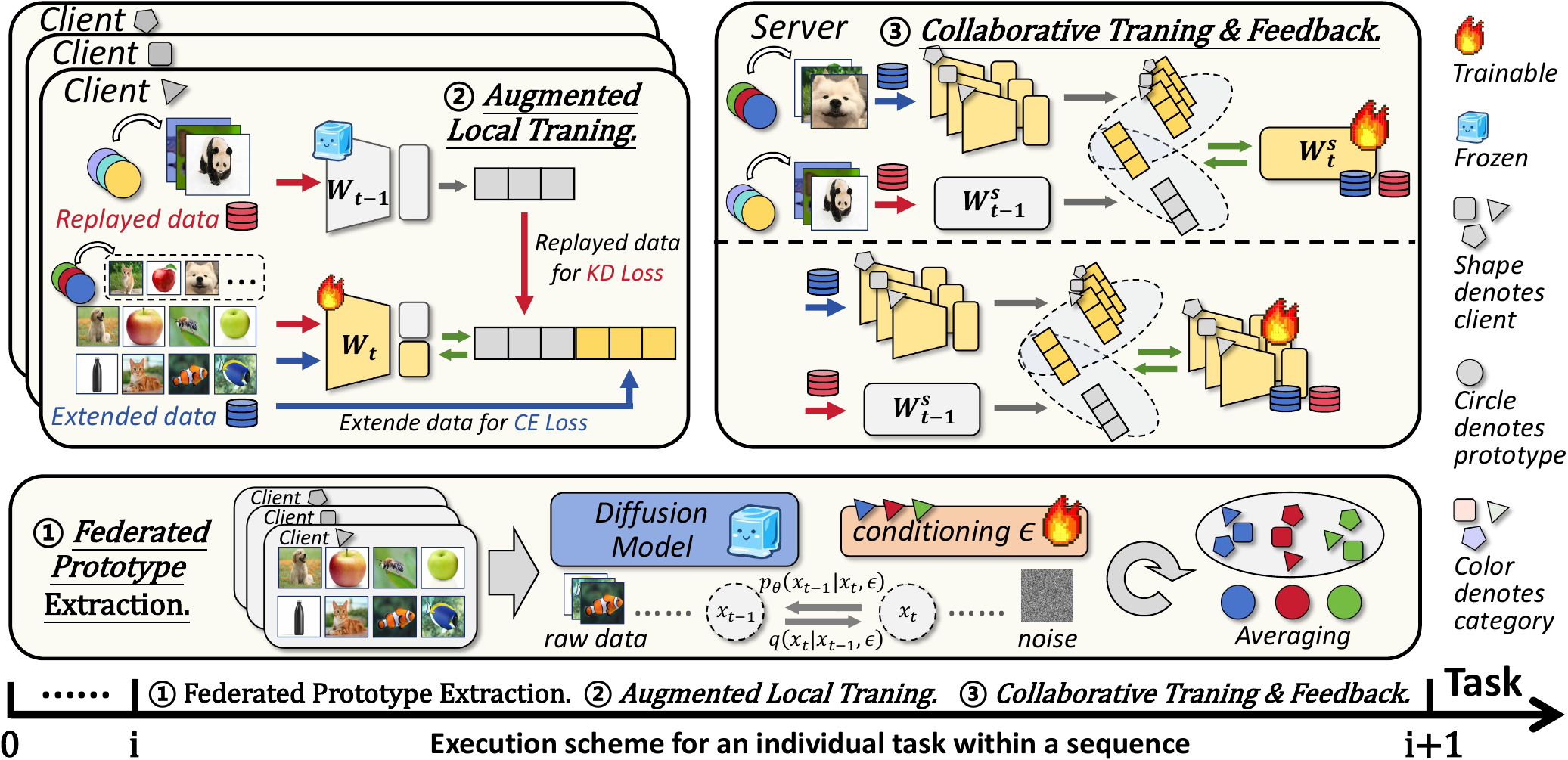}\\
    \end{minipage}
    \caption{The framework of \modelname:  
    \ding{172} \textit{Federated Prototype Extraction.} Clients extract class prototypes using a frozen pre-trained diffusion model. These prototypes enable the generation of synthetic data aligned with the current task’s knowledge, as well as the replay of previously learned knowledge. 
    \ding{173} \textit{Augmented Local Training.} On the client side, synthetic data that conform to the current task’s data distribution and preserve knowledge from previous tasks are mixed with local real data to perform local training. 
    \ding{174} \textit{Collaborative Traning \& Feedback.} On the server side, knowledge is transferred from clients’ heterogeneous models and the server’s past-task models to the current global model via the synthetic dataset; simultaneously, feedback is sent back to refine the local client models.}
    \label{fig:model_architecture}
 \end{figure*}

\section{Problem Formulation}
In this paper, we aim to continually train a server-side model through collaborative learning with heterogeneous device models over a sequence of tasks in a data-free manner. 
Specifically, we consider a federated learning system consisting of one central server and \(K\) heterogeneous clients, indexed by \( k = 1, \dots, K \).
Each client \( k \) possesses a local model \( \mathcal{M}_k \) , whose architecture may differ from others in depth, width, or structural design.
Over time, all clients sequentially encounter a class-incremental task flow \( \{t\}_{t=1}^T \), where each task \( t \) introduces a new set of classes that are strictly disjoint from those in all previous and future tasks. 
Within each task \( t \), client \( k \) holds a private local dataset \(\mathcal{D}_k^{(t)} = \{(x_{i}^{(t)}, y_{i}^{(t)})\}_{i=1}^{N_{k}^{(t)}} \), where \( N_{k}^{(t)} \) denotes the number of samples available to client \( k \) at task \( t \). The data distributions \( \{\mathcal{D}_k^{(t)}\}_{k=1}^K \) are typically non-IID across clients. 

The objective is to effectively train the server model \( \mathcal{M}_s \) using heterogeneous models, enabling it to accumulate knowledge across all tasks while avoiding catastrophic forgetting.
This could be formalized as: 
\begin{equation}
\min_{\mathcal{M}_s}  \sum_{k=1}^{K} \mathcal{L}_{CE}(\mathcal{M}_s; \mathcal{D}) 
\end{equation}
where learned knowledge \( \mathcal{D} = \bigcup_{t=1}^{T} \bigcup_{k=1}^{K} D_{k}^{(t)}\), \( D_{k}^{(t)} \) represents the dataset of client \( k \) for task \( t \) and \(\mathcal{L}( \cdot )\) denotes the loss function.

\section{METHODOLOGY}

As shown in Figure.\ref{fig:model_architecture}, we propose a model-heterogeneous federated continual learning framework, \modelname. 
First, we design the \textit{federated prototype extraction} strategy to enable task-adaptive data synthesis (see Sec.\ref{Methodology_moudle_1}). This module dynamically aligns knowledge with the current task while preserving and replaying previously learned concepts.
Then, in the \textit{augmented local training} phase (Sec.\ref{Methodology_moudle_2}), each client combines synthetic data with its private real data, where the synthetic data serve two purposes: (i) replaying knowledge from past tasks to mitigate catastrophic forgetting, and (ii) augmenting current-task coverage to alleviate data scarcity and Non-IID bias. This enables robust and forgetting-resistant local model updates. 
Finally, during \textit{collaborative training and feedback} (Sec.\ref{Methodology_moudle_3}), the server aggregates knowledge from heterogeneous client models and historical checkpoints via the synthetic dataset; simultaneously, this distilled knowledge is fed back to clients to guide client-side model refinement. Further details can be seen in Algorithm.\ref{Alog:overall}.

\subsection{Federated Prototype Extraction.}
\label{Methodology_moudle_1}

Given the continuous changes in the target domain across tasks, it is crucial to dynamically align with emerging knowledge and save replayable representations in a storage-efficient manner.
Inspired by class inversion  based on pre-trained diffusion models\cite{textual_inversion}  and its recent success in the federated setting\cite{dddr}\cite{yang2024exploring}\cite{yang2024feddeo}, we further investigate its potential in the model-heterogeneous federated continual learning, a largely unexplored yet critical setting. Building upon this technique, we implement an effective knowledge extraction mechanism to efficiently align and preserve category-level information across the continual task flow.

\textbf{Diffusion Process.}
Diffusion models are a class of generative models that learn to synthesize data by reversing a gradual noising process\cite{ddpm}\cite{ddim}. 
We denote \( p_{\text{data}} \) as the raw data distribution and \( p_{\text{latent}} = \mathcal{N}(\mathbf{0}, \mathbf{I}) \) as the target latent distribution, where \( \mathcal{N} \) is the Gaussian distribution.
The forward diffusion process follows a fixed Markov chain that progressively adds noise to original distribution  \(p_{\text{data}}\) until it converges to \( p_{\text{latent}} \). 
During the reverse denoising process, the model is trained to iteratively remove the noise to construct the desired data. 
Formally, let \( x_0 \sim p_{\text{data}} \). The whole forward process can be formulated as:

\begin{equation}
\label{eq:diffusion_forward}
\begin{aligned}
    q\left(x_{1:T} \mid x_0\right)  &:= \prod_{t=1}^T q\left(x_t \mid x_{t-1}\right), \ \text{where} \\
     q\left( {x}_t \mid {{x}_{{t - 1}}} \right) & := {\cal{N}}\left( {{x}_t ;\sqrt {{\alpha_t }}{{x}_{{t - 1}}},{{\left( {1 - {\alpha_t }} \right)}}{\bf{I}}} \right),
\end{aligned}
\end{equation}
the \(x_t\) is the noisy sample at timestep \(t\) and hyperparameters \(\alpha_t\in (0,1)\) are some pre-defined, gradually increasing positive constants that control the scale of injected noise.

Subsequently, we perform stepwise denoising on the noisy data. 
The reverse diffusion process supports conditioning. Let \( x_T \sim p_{\text{latent}} \). The reverse process from \( x_T \) to clean \( x_0  \) can be formulated as:

\begin{equation}
\begin{aligned}
    p_\theta\left(\mathbf{x}_{0: T},\mathcal{C}\right):=p\left(\mathbf{x}_T\right) \prod_{t=1}^T p_\theta\left(\mathbf{x}_{t-1} \mid \mathbf{x}_t ,\mathcal{C}\right), \ \text{where}\\
     p_\theta\left(\mathbf{x}_{t-1} \mid \mathbf{x}_t,\mathcal{C}\right):=\mathcal{N}\left(\mathbf{x}_{t-1} ; \boldsymbol{\mu}_\theta\left(\mathbf{x}_t, t, \mathcal{C}\right), \sigma_t^2 \mathbf{I}\right),  
\end{aligned}
\end{equation}
the $\mu_{\theta}$ is a neural network that predicts the mean, and $\sigma_t^2$ is a constant that varies with time.

\textbf{Prototype Training.}
In this work, we utilize a pre-trained diffusion model with its noise prediction network parameter fixed, and initializes a lightweight, learnable prototype embedding \(p\) for every new category encountered in the continual task flow. With these trainable prototypes, the client avoids training a generative model from scratch; instead, it only needs to search for category-specific conditions that guide the generation process.
The simplified training objective becomes:

\begin{equation}
\label{eq:proto_loss}
\mathcal{L}_{\text{proto}}(p) = \mathbb{E}_{t, x_0 \sim p_{\text{data}},\, \epsilon \sim \mathcal{N}(\mathbf{0}, \mathbf{I})} \left\| \epsilon - \epsilon_{\theta}(x_t, t, p) \right\|_2^2 ,
\end{equation}
where the prediction network  \(\epsilon_\theta(\cdot)\)  is kept frozen during training,  and only the prototype \( p \) is optimized to minimize the discrepancy between predicted and true noise \(\epsilon\).

After local training, each client uploads its category-specific prototype \( p_{i,k} \)  (where \(i\) denotes the category index and \(k\) the client index) to the server. The server then aggregates these local prototypes across all participating clients to construct a federated prototype \( p_k \) as follows:

\begin{equation}
p_i = \frac{1}{K} \sum_{k=1}^{K} p_{i,k}.
\end{equation}

Building upon the globally informed prototype, we can dynamically matching the current knowledge and further utilize them in downstream steps to: (1) enhance model generalization and robustness in non-IID scenarios, (2) serve as a proxy dataset for knowledge distillation in heterogeneous models and (3) combat catastrophic forgetting by facilitating effective cross-task knowledge transfer.

\subsection{Augmented Local Continual Training.}
\label{Methodology_moudle_2}

When a client encounters a new task in the continual task flow, local training faces two fundamental challenges: (i) intra-task Non-IID bias, where the local data distribution may deviate from the global population, leading to suboptimal and unstable updates; and (ii) inter-task catastrophic forgetting, where learning the current task overwrites knowledge from previously learned tasks due to the absence of historical data.
Leveraging the prototypes obtained, we can enhance local client updates through data augmentation. In this phase, each client synthesizes category-conditioned samples using the aggregated federated prototypes and blends them with its private real data. This hybrid training strategy not only enriches the local data distribution to counter Non-IID bias and scarcity, but also replays past task knowledge in a storage-efficient manner, enabling continual learning without explicit data retention. 

\textbf{Adaptive Classifier Expansion.}
To preserve accumulated knowledge while adapting to evolving task environments in a continual learning setting, we dynamically expand the model’s classification head capacity for novel classes introduced by incoming tasks.
Formally, let the model at task \(t\) to be denoted as \( \mathcal{M}^{(t)} = \mathcal{M}(\theta \diamond \theta^{(t)}) \), comprising a shared feature extractor a shared feature extractor \( \theta \) (which may be frozen or fine-tuned) and a task-adaptive classification head \( \theta^{(t)} \). 
When transitioning from task \( t-1 \) to \( t \), the classification head \( \theta^{(t)} \) is adaptively extended by retaining the parameters \( \theta^{(t-1)} \) responsible for previously learned classes, while introducing a newly initialized parameter block \( \theta^{(t),\text{new}} \) for incoming classes. 

By explicitly separating old and new knowledge at the classification head level, we enable more effective loss function design tailored to each knowledge type, facilitating the preservation of previously learned concepts while promoting efficient assimilation of new ones.

\textbf{Knowledge Replay for Inter-Task Forgetting.}
When the model is assimilating new knowledge from task \( t \), it is equally crucial to recall the knowledge accumulated from previous tasks \( \{0, 1, \ldots, t-1\} \) to prevent inter-task catastrophic forgetting and maintain continuity. 
Since the stored prototypes capture knowledge from past tasks, we leverage them to condition a pre-trained diffusion model and synthesize samples for effective konwledge replay without accessing original data. 
Specifically, during local training at each client, we generate a set of synthetic data \( \mathcal{D}^{\text{syn}}_{\text{pre}}\) representing previously learned classes. These labeled synthetic samples can be directly used for direct classification supervision via standard cross-entropy loss.
Then, we adopt the previous model \( \mathcal{M}_k^{(t-1)}\) as the teacher model for knowledge distillation. We define the inter-task knowledge preservation loss for client \(k\) as:

\begin{equation}
\begin{aligned}
\mathcal{L}_{k}^{(t),\text{inter}} & = \mathcal{L}_{ce}(\mathcal{D}^{\text{syn}}_{\text{pre}};\mathcal{M}_k(\theta \diamond \theta^{(t-1)})) 
\\
& + \mathcal{L}_{kl} (\mathcal{D}^{\text{syn}}_{\text{pre}};\mathcal{M}_k(\theta \diamond \theta^{(t-1)}) , \mathcal{M}_k^{(t-1)} ),
\label{eq:client_loss_inter}
\end{aligned}
\end{equation}
where \( \mathcal{L}_{ce}(\cdot) \) is the standard cross-entropy loss using hard labels, \( \mathcal{L}_{kl}(\cdot) \) computes Kullback-Leibler divergence between the student’s and teacher’s output distributions. Note that this loss function targets the client model component \(\mathcal{M}_k(\theta \diamond \theta^{(t-1)})\) responsible for preserving old knowledge

\textbf{Data Augmentation for Intra-Task Bias.}
Within a given task \(t\), local clients often exhibit heterogeneous data distributions. This heterogeneity can lead to divergent model updates, unstable convergence and degraded model performance. 

To mitigate this issue, each client generates synthetic samples \( \mathcal{D}^{\text{syn}}_{\text{cur}}\) pre-trained diffusion model on the current task’s federated prototypes. These synthetic samples represent an unbiased, global knowledge representation of the current task, and are subsequently combined with the client’s private real data \( \mathcal{D}^{(t)}_{k}\) to form an augmented training set. The component \(\mathcal{M}(\theta \diamond \theta^{(t),\text{new}})\), responsible for learning incoming knowledge, is optimized using the standard cross-entropy loss over this hybrid dataset:

\begin{equation}
\mathcal{L}_{k}^{(t),\text{intra}} = \mathcal{L}_{ce}(\mathcal{D}^{\text{syn}}_{\text{cur}} \cup \mathcal{D}^{(t)}_{k};\mathcal{M}_k(\theta \diamond \theta^{(t),\text{new}})).
\label{eq:client_loss_intra}
\end{equation}

By integrating both the inter-task knowledge preservation loss $\mathcal{L}_{k}^{(t),\text{inter}}$ (Eq.~\ref{eq:client_loss_inter}) and the intra-task bias mitigation loss $\mathcal{L}_{k}^{(t),\text{intra}}$ (Eq.~\ref{eq:client_loss_intra}), our framework enables augmented local training in continual federated settings:
\begin{equation}
\mathcal{L}_{k}^{(t)} = \mathcal{L}_{k}^{(t),\text{inter}} + \mathcal{L}_{k}^{(t),\text{intra}}.
\label{eq:client_loss_tot}
\end{equation}

This dual-loss design ensures that local updates achieve effective retention of old knowledge and stable assimilation of new knowledge.

\subsection{Collaborative Distillation and Feedback.}
\label{Methodology_moudle_3}

Given the heterogeneous models across clients, it is essential to enable cross-model knowledge communication to facilitate cross-client knowledge transfer and collaboratively train an effective global server model.
In this phase, we design a collaborative mechanism that consolidates distributed knowledge from heterogeneous client models into the server model, enabling the server to continuously distill and accumulate knowledge from diverse clients throughout the continual task stream.

\textbf{Knowledge Transfer for Server.}
We aim to construct a global server model that encapsulates the comprehensive knowledge from distributed clients and the continually evolving knowledge from task flow.
Similar to the previous data synthesis procedure, we leverage federated prototypes to synthesize two types of data: (i) current-task samples \(\mathcal{D}^{\text{syn'}}_{\text{cur}}\), representing semantic concepts of the latest task; and (ii) historical-task samples \(\mathcal{D}^{\text{syn'}}_{\text{pre}}\), preserving knowledge from previously learned tasks. These synthetic datasets enable data-free knowledge distillation without accessing any public dataset.

To transfer the knowledge of the latest task from distributed heterogeneous clients to the global server model, we distill the collective intelligence of client models by aligning the server’s predictions with the ensemble knowledge on \(\mathcal{D}^{\text{syn'}}_{\text{cur}}\). Specifically, we define the current-task distillation loss as:

\begin{equation}
\begin{aligned}
\mathcal{L}_{s}^{(t),\text{cur}} = \mathcal{L}_{kl} (\mathcal{D}^{\text{syn}'}_{\text{cur}};\mathcal{M}_s(\theta \diamond \theta^{(t),\text{new}}) , \frac{1}{K} \sum_{k=1}^{K} \mathcal{M}_k^{(t)} ),
\label{eq:server_loss_cur}
\end{aligned}
\end{equation}
where \(\mathcal{M}_s(\theta \diamond \theta^{(t),\text{new}})\) is the server model with classification head for new classes and \(\frac{1}{K} \sum_{k=1}^{K} \mathcal{M}_k^{(t)}\) represents the averaged prediction over \(K\) participating clients. 

Subsequently, to preserve knowledge from previous tasks, we employ the server model from the last task, \(\mathcal{M}_s^{(t-1)}\),as a teacher and distill its predictions over \(\mathcal{D}^{\text{syn}'}_{\text{pre}}\) into the current server model. This yields the knowledge preservation loss for previous knowledge:
\begin{equation}
\begin{aligned}
\mathcal{L}_{s}^{(t),\text{pre}} = \mathcal{L}_{kl} (\mathcal{D}^{\text{syn}'}_{\text{pre}};\mathcal{M}_s(\theta \diamond \theta^{(t-1)}) , \mathcal{M}_s^{(t-1)} ),
\label{eq:server_loss_pre}
\end{aligned}
\end{equation}
where \(\mathcal{D}^{\text{syn}'}_{\text{pre}}\) denotes the server’s model with classifier part for old classes. By jointly optimizing new knowledge assimilation and old knowledge preservation through synthetic data distillation, the server evolves to effectively absorb emerging task knowledge while retaining previously learned concepts. The total server-side training objective is formulated as:
\begin{equation}
\mathcal{L}_{k\rightarrow s}^{(t)} = \mathcal{L}_{s}^{(t),\text{cur}} + \mathcal{L}_{s}^{(t),\text{pre}}.
\label{eq:server_loss_tot}
\end{equation}

\textbf{Knowledge Feedback for Clients.} 
To propagate consolidated global knowledge back to clients, we
adapt the server-side distillation objective by replacing the server model \(\mathcal{M}_s\) with \(\mathcal{M}_k\). Since server model acts as a global knowledge integrator that maintains comprehensive awareness of all previously learned tasks, this distillation mechanism enables each client to align its representation with both the current-task ensemble and the historically consolidated server knowledge. Thus, we define the loss function as:

\begin{equation}
\begin{aligned}
\mathcal{L}_{s\rightarrow k}^{(t)} &= 
\mathcal{L}_{kl} (\mathcal{D}^{\text{syn}'}_{\text{cur}};\mathcal{M}_k(\theta \diamond \theta^{(t),\text{new}}) , \frac{1}{K} \sum_{k=1}^{K} \mathcal{M}_k^{(t)} )
\\
& + \mathcal{L}_{kl} (\mathcal{D}^{\text{syn}'}_{\text{pre}};\mathcal{M}_k(\theta \diamond \theta^{(t-1)}) , \mathcal{M}_s^{(t-1)} ).
\label{eq:server_loss_tot2}
\end{aligned}
\end{equation}

In the preceding paragraphs, we have presented the three core steps of \modelname \ for federated ensemble knowledge transfer under heterogeneous client models trained on evolving task streams. The complete training details can be seen in Algorithm.\ref{Alog:overall}.

\begin{algorithm}[t]
\DontPrintSemicolon
\SetAlCapHSkip{0ex} 
\caption{Illustration of \modelname}
\label{Alog:overall}  
\kwInit{initial server model \(\mathcal{M}_s\), client models \(\{\mathcal{M}_k\}\), client number \(K\), task number \(T\), and number of rounds \(Q, Q_{k \rightarrow s}, Q_{s \rightarrow k}, Q_p, Q_l\).}
\SetKwFunction{FMain}{\modelname}
\SetKwFunction{CU}{ClientUpdate}
\SetKwFunction{PG}{PrototypeUpdate}
\SetKwProg{Fn}{Function}{:}{}
\SetKwProg{FnMain}{Algorithm}{:}{}
\SetKw{KwOutput}{Output:}
\FnMain{\FMain}{
    \For{task \(t \ = 1 \ to \ T\)}{
    Execute \PG{} on all clients \\
    }
    \For{communication round \(q \ = 1 \ to \ Q\)}{
        Execute \CU{} on all clients \\
        Generate \(\mathcal{D}^{\text{syn'}}_{\text{cur}}\) and \(\mathcal{D}^{\text{syn'}}_{\text{pre}}\) with \(\{p_i\}\) \\
        \For{server distillation round \(q \ = 1 \ to \ Q_{k \rightarrow s}\)}{
            Update server model \(\mathcal{M}_s\) via Eq.\ref{eq:server_loss_tot} \\
        }
        \For{client distillation round \(q \ = 1 \ to \ Q_{s \rightarrow k}\)}{
            Update \(\mathcal{M}_k\) via Eq.\ref{eq:server_loss_tot} for each client \(k\) \\
        } 
    }
}
\Fn{\PG}{
    Initialize prototype \( p_i\) for new class \(i\) \\
    \For{prototype generation round \(q \ = 1 \ to \ Q_p\)}{
        \For{client \(k \ = 1 \ to \ K\)}{
            Update \(p_i\) on private data via Eq.\ref{eq:proto_loss} \\
        }
        \(p_i = \frac{1}{K} \sum_{k=1}^{K} p_{i,k}\) \\
    }
}
\Fn{\CU}{
    Generate \(\mathcal{D}^{\text{syn}}_{\text{cur}}\) and \(\mathcal{D}^{\text{syn}}_{\text{pre}}\) with \(\{p_i\}\) \\
        \For{local round \(q \ = 1 \ to \ Q_l\)}{
            Compute the local gradient \(\mathcal{L}_{k}^{(t)}\) via Eq.\ref{eq:client_loss_tot} \\
            \(\mathcal{M}_{s,q+1}^{(t)} \leftarrow \mathcal{M}_{s,q}^{(t)} - \gamma \nabla \mathcal{L}_{k}^{(t)} \)
    }}
\end{algorithm}

\section{Evaluation}

\begin{table*}[t]
\caption{Performance comparison on cumulative tasks after learning each task (\(t_1\)\(\to\)\(t_2\)\(\to\)\(t_3\)) under different Dirichlet distributions. A dash ('–') indicates that the method is not applicable to the given experimental setting. The lightgray highlights the metrics evaluated after all tasks completed. The bold is the optimal result and the underline is the second best.}
\label{tab:overall}
\centering
\renewcommand{\arraystretch}{1.1}
\tabcolsep=4.8pt
\begin{tabular}{cccccccccccccccccccc}
\toprule
\multirow{3}{*}{Algorithm} & \multicolumn{9}{c}{\textit{Grayscale}} &  & \multicolumn{9}{c}{\textit{RGB}} \\ \cmidrule{2-10}  \cmidrule{12-20}

 & \multicolumn{4}{c}{\(Dir(\gamma = 0.3)\)} &  & \multicolumn{4}{c}{\(Dir(\gamma = 0.5)\)} &  & \multicolumn{4}{c}{\(Dir(\gamma = 0.3)\)} &  & \multicolumn{4}{c}{\(Dir(\gamma = 0.5)\)} \\ \cmidrule{2-5} \cmidrule{7-10} \cmidrule{12-15} \cmidrule{17-20} 
 
 & -\(t_1\) & -\(t_2\) & \cellcolor{lightgray!30}-\(t_3\) \(\uparrow\) & \cellcolor{lightgray!30}F \(\downarrow\) &  & -\(t_1\) & -\(t_2\) & \cellcolor{lightgray!30}-\(t_3\) \(\uparrow\) & \cellcolor{lightgray!30}F \(\downarrow\) &  & -\(t_1\) & -\(t_2\) & \cellcolor{lightgray!30}-\(t_3\) \(\uparrow\) & \cellcolor{lightgray!30}F \(\downarrow\) &  & -\(t_1\) & -\(t_2\) & \cellcolor{lightgray!30}-\(t_3\) \(\uparrow\) & \cellcolor{lightgray!30}F \(\downarrow\) \\ \cmidrule{1-10} \cmidrule{12-20} 
 
\rowcolor{mypurple} \multicolumn{20}{c}{\textit{\textcolor{gray}{federated continual learning with homogeneous models}}}\\

Fedavg & 93.06 & 46.40 & \cellcolor{lightgray!30}26.61 & \cellcolor{lightgray!30}92.92
 &  & 94.98 & 46.99 & \cellcolor{lightgray!30}26.74 & \cellcolor{lightgray!30}94.47
 &  & 73.00 & 37.15 & \cellcolor{lightgray!30}27.20 & \cellcolor{lightgray!30}73.65
 &  & 73.90 & 38.30 & \cellcolor{lightgray!30}27.50 & \cellcolor{lightgray!30}75.25 \\ 
FedEWC & 94.04 & 76.65 & \cellcolor{lightgray!30}61.33 & \cellcolor{lightgray!30}38.58
 &  & 95.44 & 77.66 & \cellcolor{lightgray!30}56.77 & \cellcolor{lightgray!30}49.20
 &  & 75.80 & 39.55 & \cellcolor{lightgray!30}31.97 & \cellcolor{lightgray!30}66.35
 &  & 75.10 & 38.75 & \cellcolor{lightgray!30}33.53 & \cellcolor{lightgray!30}65.35 \\ 
FedLWF-2T & 93.06 & 77.88 & \cellcolor{lightgray!30}68.69 & \cellcolor{lightgray!30}9.10
 &  & 94.98 & 82.1 & \cellcolor{lightgray!30}72.97 & \cellcolor{lightgray!30}13.54
 &  & 73.00 & 52.65 & \cellcolor{lightgray!30}48.57 & \cellcolor{lightgray!30}20.20
 &  & 73.90 & 55.50 & \cellcolor{lightgray!30}49.23 & \cellcolor{lightgray!30}23.60 \\ 
Target & 93.06 & 78.75 & \cellcolor{lightgray!30}71.92 & \cellcolor{lightgray!30}10.48
 &  & 94.98 & 84.89 & \cellcolor{lightgray!30}76.51 & \cellcolor{lightgray!30}11.52
 &  & 73.00 & 52.55 & \cellcolor{lightgray!30}43.50 & \cellcolor{lightgray!30}29.50
 &  & 73.90 & 53.65 & \cellcolor{lightgray!30}48.50 & \cellcolor{lightgray!30}28.80 \\ 
Lander & - & - & \cellcolor{lightgray!30}- & \cellcolor{lightgray!30}-
 &  & - & - & \cellcolor{lightgray!30}- & \cellcolor{lightgray!30}-
 &  & 74.90 & 57.45 & \cellcolor{lightgray!30}50.43 & \cellcolor{lightgray!30}16.55
 &  & 77.8 & 61.20 & \cellcolor{lightgray!30}57.03 & \cellcolor{lightgray!30}12.80 \\ 
\midrule
\rowcolor{mypurple} \multicolumn{20}{c}{\textit{\textcolor{gray}{federated continual learning with heterogeneous models}}}\\
FT + KD & 86.58 & 43.22 & \cellcolor{lightgray!30}17.29 & \cellcolor{lightgray!30}86.50
 &  & 91.54 & 45.46 & \cellcolor{lightgray!30}16.27 & \cellcolor{lightgray!30}91.22
 &  & 63.10 & 33.95 & \cellcolor{lightgray!30}25.27 & \cellcolor{lightgray!30}65.50 
 &  & 74.00 & 43.30 & \cellcolor{lightgray!30}28.40 & \cellcolor{lightgray!30}80.30 \\ 
EWC + KD & 85.64 & 29.78 & \cellcolor{lightgray!30}15.86 & \cellcolor{lightgray!30}71.26 
 &  & 92.94 & 31.56 & \cellcolor{lightgray!30}16.75 & \cellcolor{lightgray!30}76.07 
 &  & 67.00 & 33.60 & \cellcolor{lightgray!30}27.33 & \cellcolor{lightgray!30}67.10 
 &  & 75.60 & 41.65 & \cellcolor{lightgray!30}28.97 & \cellcolor{lightgray!30}79.45 \\ 
LWF + KD & 86.58 & 67.37 & \cellcolor{lightgray!30}61.97 & \cellcolor{lightgray!30}12.21 
 &  & 91.54 & 79.40 & \cellcolor{lightgray!30}65.81 & \cellcolor{lightgray!30}14.85
 &  & 63.10 & 36.30 & \cellcolor{lightgray!30}33.63 & \cellcolor{lightgray!30}45.45
 &  & 74.00 & 51.15 & \cellcolor{lightgray!30}41.47 & \cellcolor{lightgray!30}59.55 \\ 
Ours & 95.32 & 87.28 & \cellcolor{lightgray!30}\textbf{80.92} & \cellcolor{lightgray!30}\textbf{4.37} 
 &  & 96.02 & 90.41 & \cellcolor{lightgray!30}\textbf{82.99} & \cellcolor{lightgray!30}\textbf{4.86}
 &  & 86.50 & 80.25 & \cellcolor{lightgray!30}\textbf{74.10} & \cellcolor{lightgray!30}\textbf{8.40} 
 &  & 88.40 & 78.50 & \cellcolor{lightgray!30}\textbf{73.43} & \cellcolor{lightgray!30}\textbf{6.85} \\ 
\bottomrule
\end{tabular}
\end{table*}

\begin{figure*}[t]
\begin{tabular}{@{\extracolsep{\fill}}c@{}c@{}c@{}c@{}@{\extracolsep{\fill}}}
\includegraphics[width=0.25\linewidth]{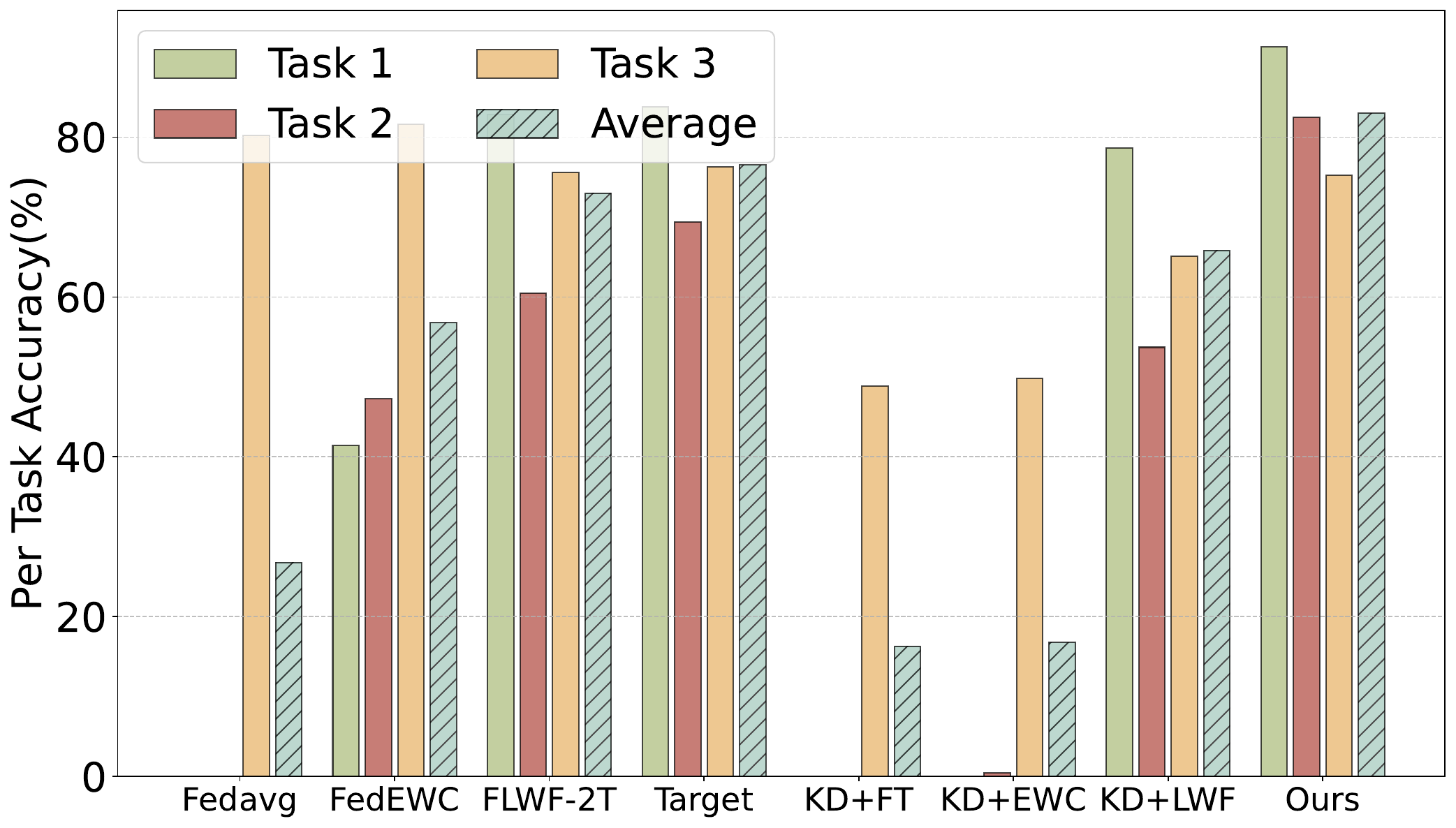}&
\includegraphics[width=0.25\linewidth]{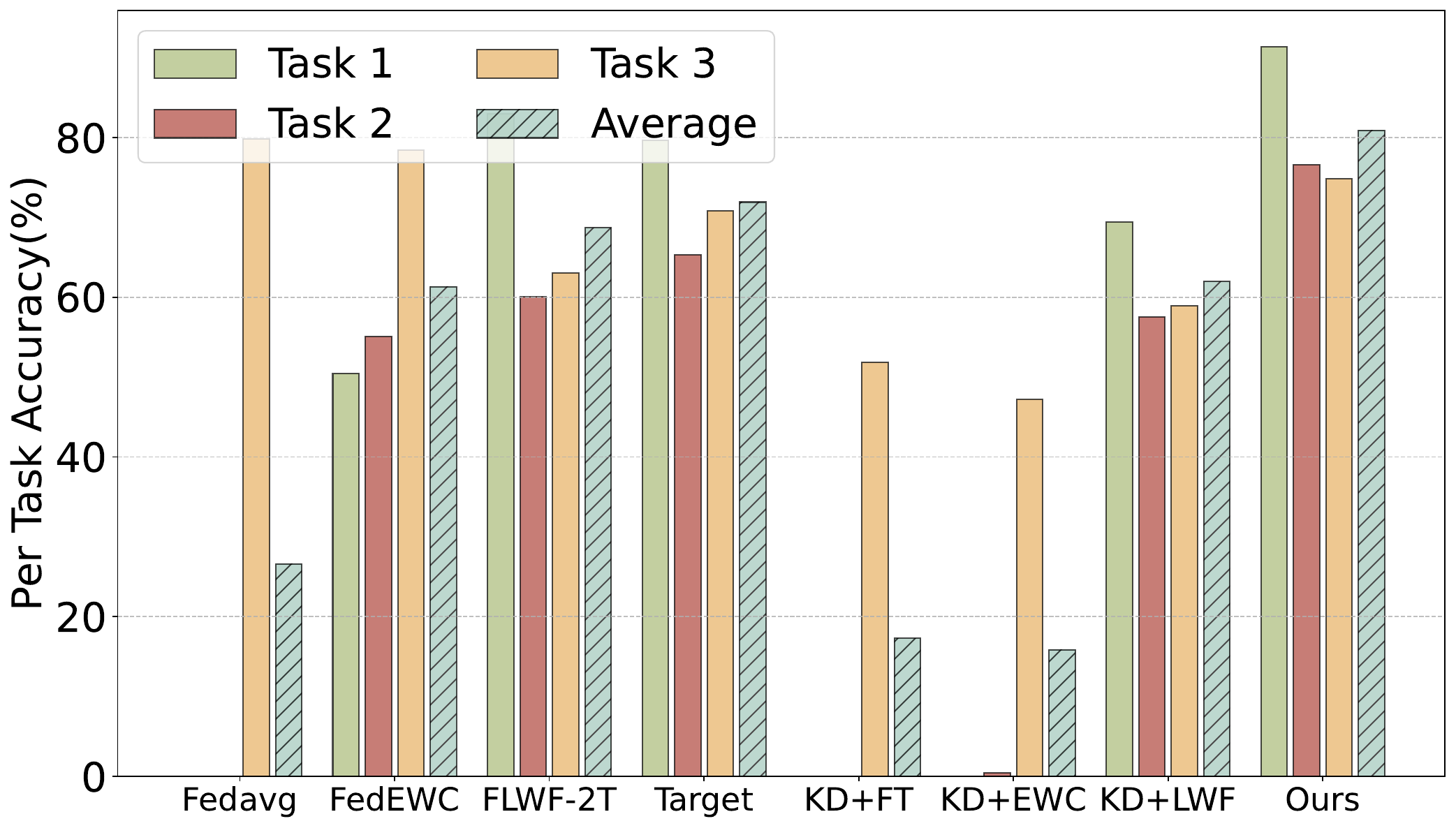}&
\includegraphics[width=0.25\linewidth]{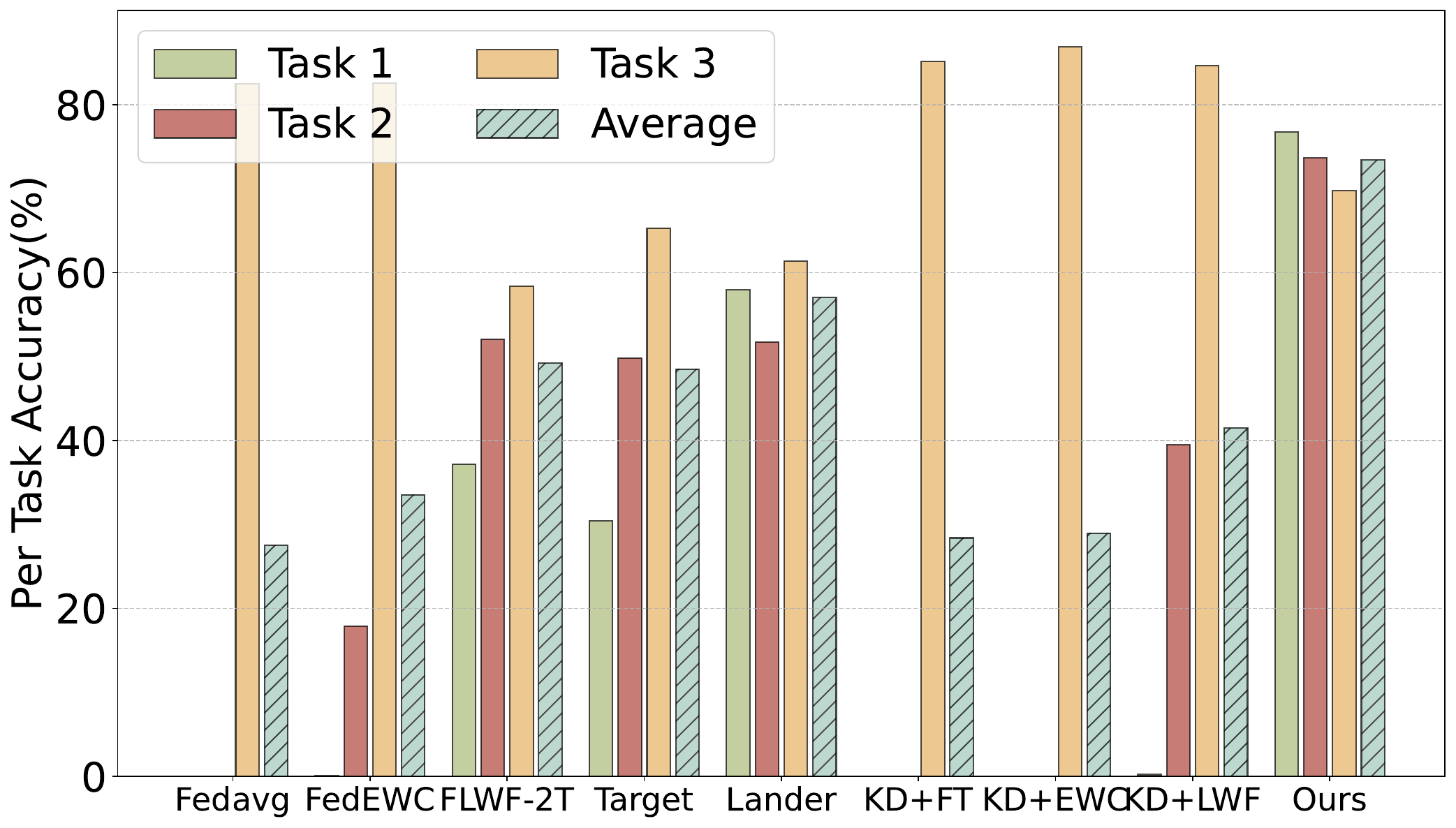}&
\includegraphics[width=0.25\linewidth]{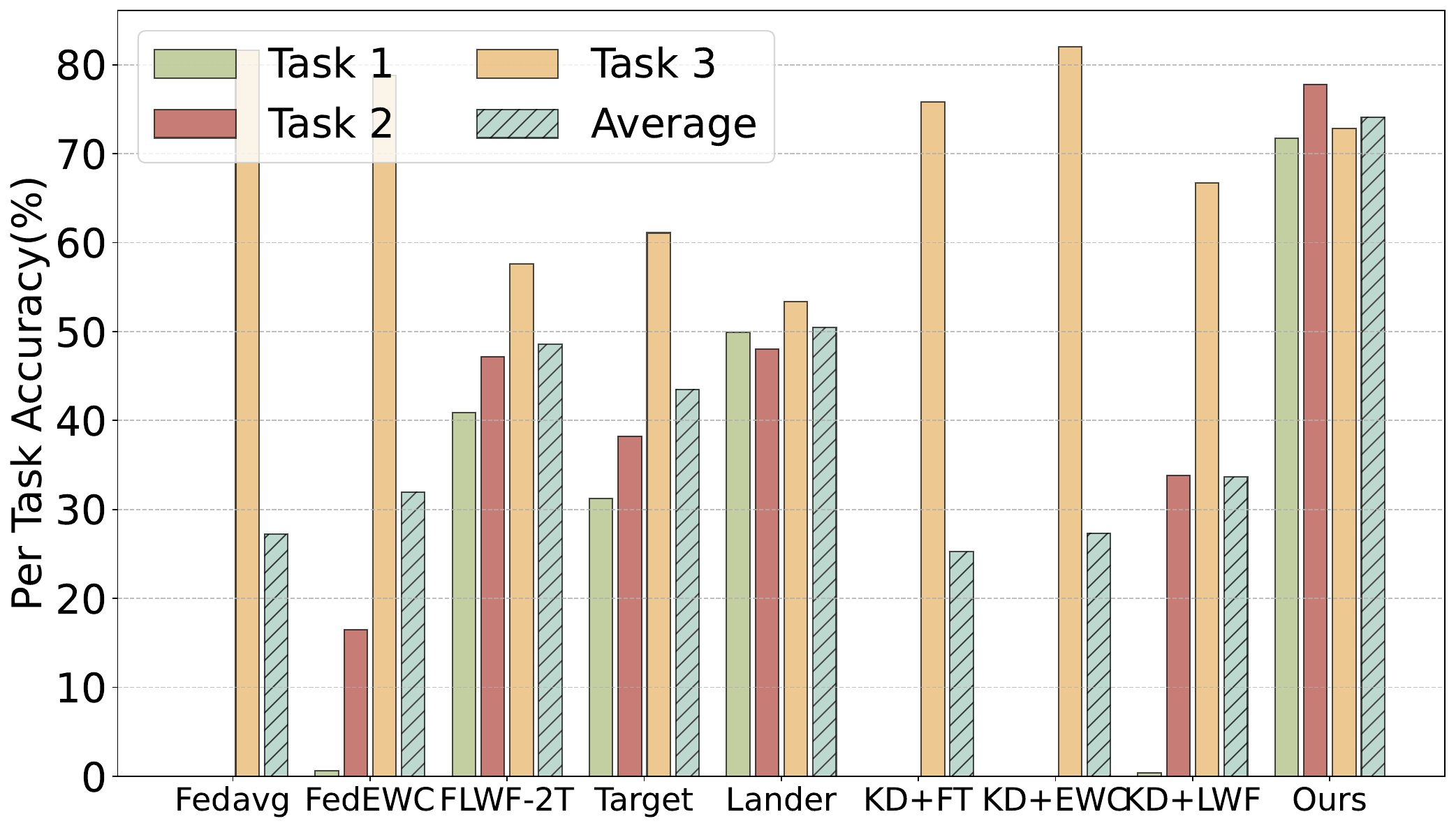}\\
(a) \textit{Grayscale} Dir($\gamma$=0.5) & (b) \textit{Grayscale} Dir($\gamma$=0.3) & (c) \textit{RGB} Dir($\gamma$=0.5) & (d) \textit{RGB} Dir($\gamma$=0.3)
\end{tabular}
\caption{Per task accuracy and average accuracy comparison of various methods. The results are evaluated after sequentially learning all tasks.}
\label{fig:performance}
\end{figure*}




\subsection{Experimental Settings}

\subsubsection{Dataset setting}
We conduct experiments on two benchmark datasets, \textit{Grayscale} and \textit{RGB}, to evaluate various methods. 
Specifically, \textit{Grayscale} benchmark is constructed by aggregating the first 10 classes from each of MNIST\cite{lecun1998gradient}, EMNIST\cite{cohen2017emnist}, and Fashion-MNIST\cite{xiao2017fashion}, forming a 30-class grayscale image classification dataset. The \textit{RGB} benchmark comprises the first 30 classes of CIFAR-100\cite{krizhevsky2009learning}, offering a 3-channel image classification dataset.

To emulate a class-incremental continual learning scenario, we divide the label space of each dataset into multiple non-overlapping tasks of equal size, where each task is learned sequentially without revisiting previous classes. Here, we set this to 3 tasks. 
Furthermore, to reflect practical data heterogeneity among clients within each task, we adopt a non-independent and identically distributed (non-IID) data partitioning scheme. The extent of data heterogeneity is described by the Dirichlet distribution $Dir(\gamma)$\cite{hsu2019measuring}, where a smaller value of $\gamma$ corresponds to a higher degree of non-IIDness. In our experiments, we evaluate two heterogeneity levels by setting $\gamma = 0.5$ and $\gamma = 0.3$, respectively. 

For fair and consistent evaluation, we utilize the original test sets corresponding to each source dataset: for \textit{Grayscale}, we combine the test sets of MNIST, EMNIST, and Fashion-MNIST; for \textit{RGB}, we use the test split of the selected 30 classes from CIFAR-100. 

\subsubsection{Model setting}
We consider a federated learning scenario in which heterogeneous client models collaboratively and continuously train a larger server-side model. Client models are constructed according to their varying resources. In our experiments, we follow the principal works in FCL\cite{target}\cite{dddr}\cite{lander}, employing a total of five clients. 
Based on their parameter scales, we categorize the client models into three types: Large (L), Medium (M), and Small (S). Client 1, possessing the abundant resources, is assigned an L-type model. Clients 2 and 3, with moderate computational capacity, are equipped with M-type models. Clients 4 and 5, facing the constrained resource budgets, are assigned S-type models. 

\begin{figure}[h]
    \centering
    \begin{minipage}[t]{1\linewidth}
    \centering
        \begin{tabular}{@{\extracolsep{\fill}}c@{}c@{}@{\extracolsep{\fill}}}
             \includegraphics[width=0.45\linewidth]{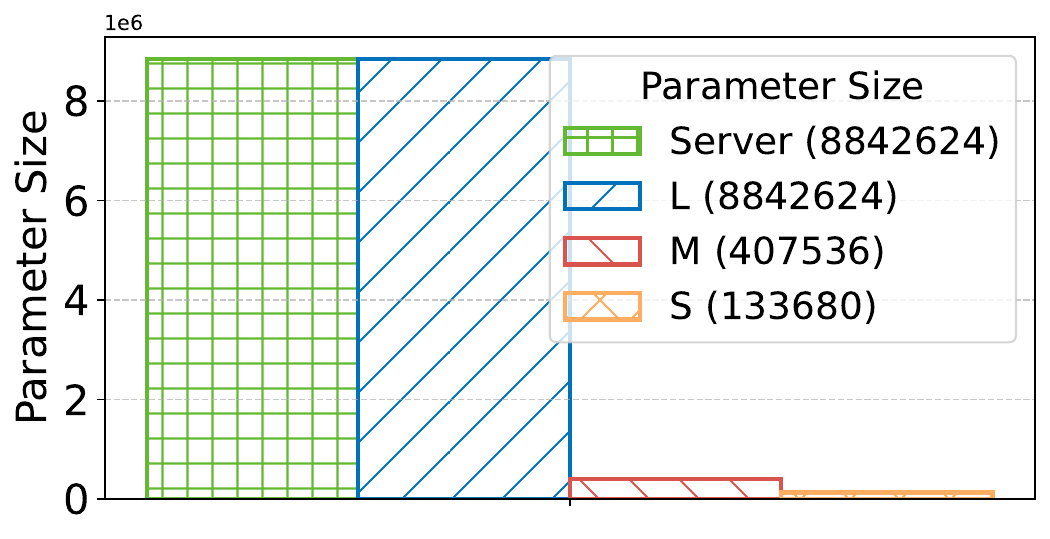} \hspace{2mm} &
            \includegraphics[width=0.45\linewidth]{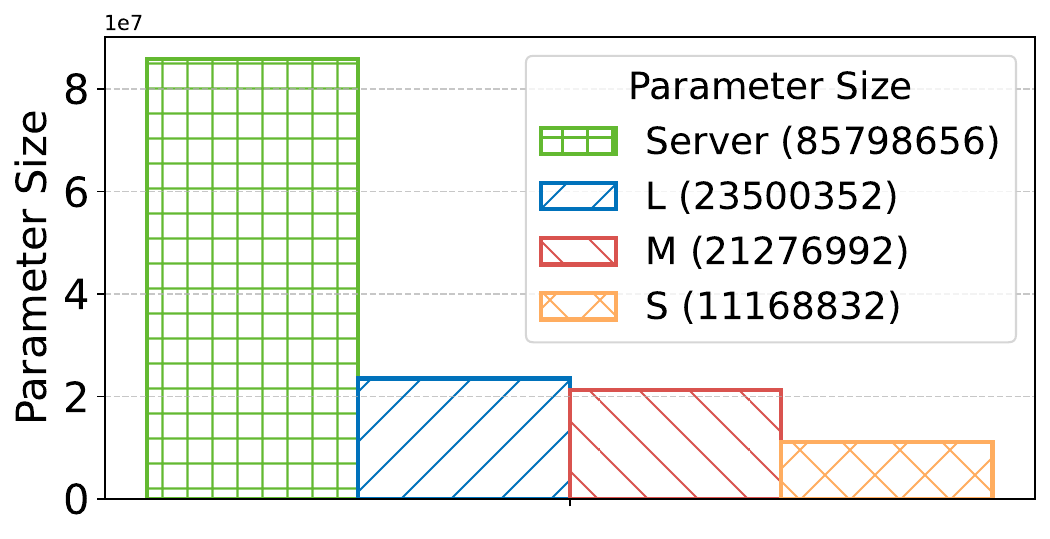}\\   
            (a) model setting for \textit{Gracyscale}. & (b) model setting for \textit{RGB}.
        \end{tabular}
    \end{minipage}
    \caption{Comparison of model parameter sizes}
    \label{fig:para_size}
 \end{figure}

For \textit{Grayscale}, a 4-layer CNN is adopted as both the L-type client model and the server model, while two 2-layer CNNs with different convolutional configurations are used as the M-type and S-type models, respectively. For \textit{RGB}, ResNet-50, ResNet-32, and ResNet-18\cite{he2016deep} are assigned as the L-type, M-type, and S-type models, respectively. On the server side, to reflect architectural diversity, we employ a pre-trained Vision Transformer (ViT)\cite{vit} and fine-tune only its classification head. The detailed parameter information is presented in Figure~\ref{fig:para_size}.

\subsubsection{Baselines}
We first consider baseline methods that assume homogeneous client models. These include the original \textit{FedAvg}\cite{Fedavg}, federated variants of classic continual learning approaches such as \textit{FedEWC}~\cite{lee2017overcoming} and \textit{FLwF-2T}\cite{fedLWF2T} inspired by LWF\cite{li2017learning}, as well as latest data-free FCL methods that rely on synthetic data, such as \textit{Target}\cite{target} and \textit{Lander}\cite{lander}. For all these homogeneous baselines, we configure the models on all five clients as S-type. Noted that Lander relies on dataset-specific pre-trained label embeddings and is therefore inapplicable to our first dataset.

 To evaluate under model heterogeneity, we further include heterogeneous-model approaches. Specifically, we adopt representative federated knowledge distillation \textit{KD} strategies\cite{fedgems}\cite{li2019fedmd} as the foundational communication mechanism for heterogeneous models, and combine them with brief fine-tuning as well as classic continual learning methods, EWC and LwF, to form additional hybrid baselines (\textit{KD + FT/EWC/LWF}). In all heterogeneous methods, the client and server model architectures are configured identically to those in our proposed framework.

\subsubsection{Evaluation metrics}
Following the previous FCL works\cite{target}\cite{dddr}\cite{lander}, we evaluate performance with two evaluation metrics: accuracy (ACC) and forgetting measure (F)\cite{forgettingmeasure}. 
specifically, we report the server's accuracy on previously accumulated tasks after each task to evaluate different methods. 
In addition, we measure forgetting after the final task. The average forgetting at k-th task can be defined as:

\begin{equation}
\begin{aligned}
    F_t & = \frac{1}{t-1} \sum_{j=1}^{t-1} f_j^t,
    \\
    f_j^t & = \max_{l \in \{1, \dots, T-1\}} acc_{l,j} - acc_{t,j}, \quad \forall j < T.
\end{aligned}
\end{equation}

It quantifies the gap between the highest accuracy ever achieved on a task throughout the learning in the past and the current retention of the model's knowledge. Lower F implies
better knowledge preservation on previous tasks.

\subsubsection{Implementation}
For \textit{Grayscale},  We configure 5 pototype update epoch and 50 local steps with 128 batch size. We set \(\mathcal{D}^{\text{syn}}_{\text{cur}}\), \(\mathcal{D}^{\text{syn}}_{\text{pre}}\), \(\mathcal{D}^{\text{syn'}}_{\text{cur}}\), \(\mathcal{D}^{\text{syn'}}_{\text{pre}}\) to 200, 100, 500 and 500, respectively. We set rounds \(Q\), \(Q_l\), \(Q_{k \rightarrow s}\), \(Q_{s \rightarrow k}\) and batch size to 5, 1, 2, 1 and 128, respectively. The pre-trained diffusion model comes from the minDiffusion~\footnote{https://github.com/byrkbrk/conditional-ddpm} trained on Mnist, Emnist and Fashion-Mnist datasets after 100 epochs.
For \textit{RGB}, We configure 10 pototype update epoch and 50 local steps with 12 batch size. 
We set \(\mathcal{D}^{\text{syn}}_{\text{cur}}\), \(\mathcal{D}^{\text{syn}}_{\text{pre}}\), \(\mathcal{D}^{\text{syn'}}_{\text{cur}}\), \(\mathcal{D}^{\text{syn'}}_{\text{pre}}\) to 200, 50, 200 and 200, respectively. We set rounds \(Q\), \(Q_l\), \(Q_{k \rightarrow s}\), \(Q_{s \rightarrow k}\) and batch size to 40, 5, 1, 1 and 128, respectively. The pre-trained diffusion model comes from the latent diffusion model (LDM) ~\footnote{https://github.com/CompVis/latent-diffusion}\cite{rombach2022highresolutionimagesynthesislatent}.

We train the model with SGD optimizer and the learning rate is 0.01. We run all experiments on the GeForce RTX 3090 GPUs 24GB.

\begin{figure}[t]
    \centering
    \begin{minipage}[t]{1\linewidth}
    \centering
        \begin{tabular}{@{\extracolsep{\fill}}c@{}c@{}@{\extracolsep{\fill}}}
             \includegraphics[width=0.43\linewidth]{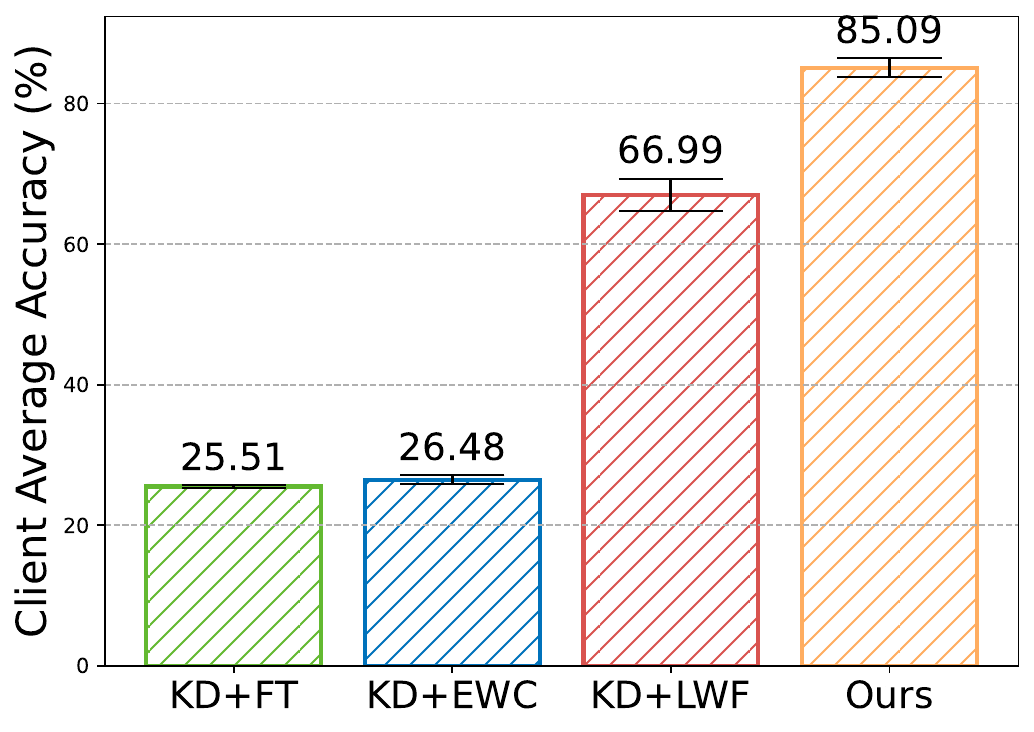} \hspace{2mm} &
            \includegraphics[width=0.43\linewidth]{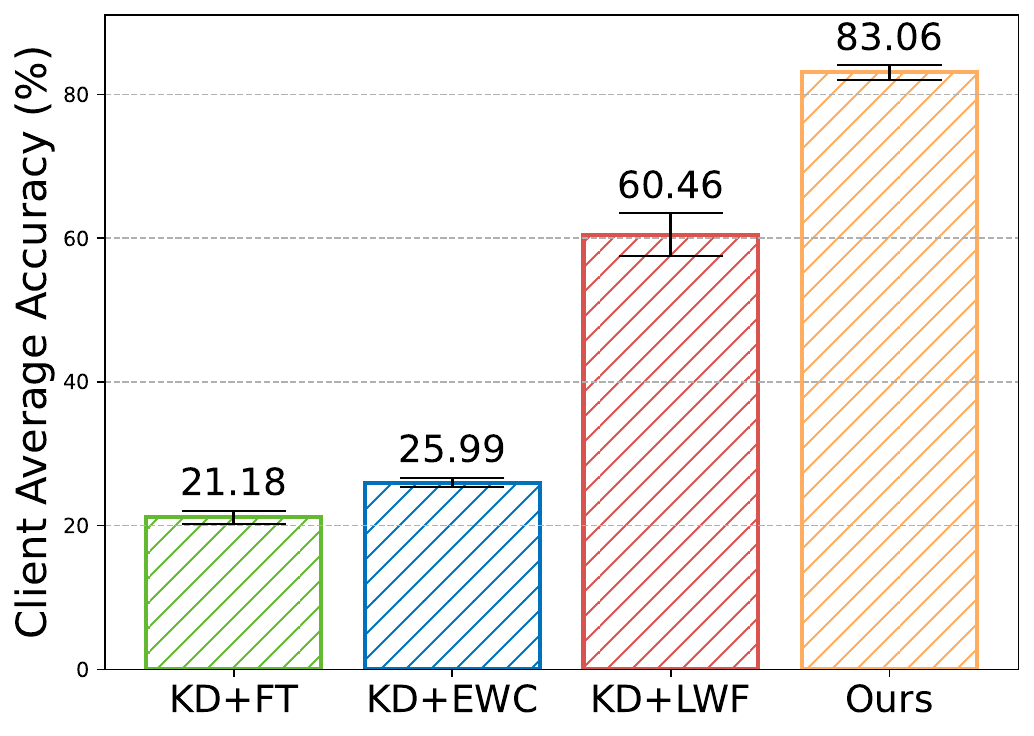}\\    
             (a) \textit{Grayscale} Dir($\gamma$=0.5) & (b) \textit{Grayscale} Dir($\gamma$=0.3) \\
             \includegraphics[width=0.43\linewidth]{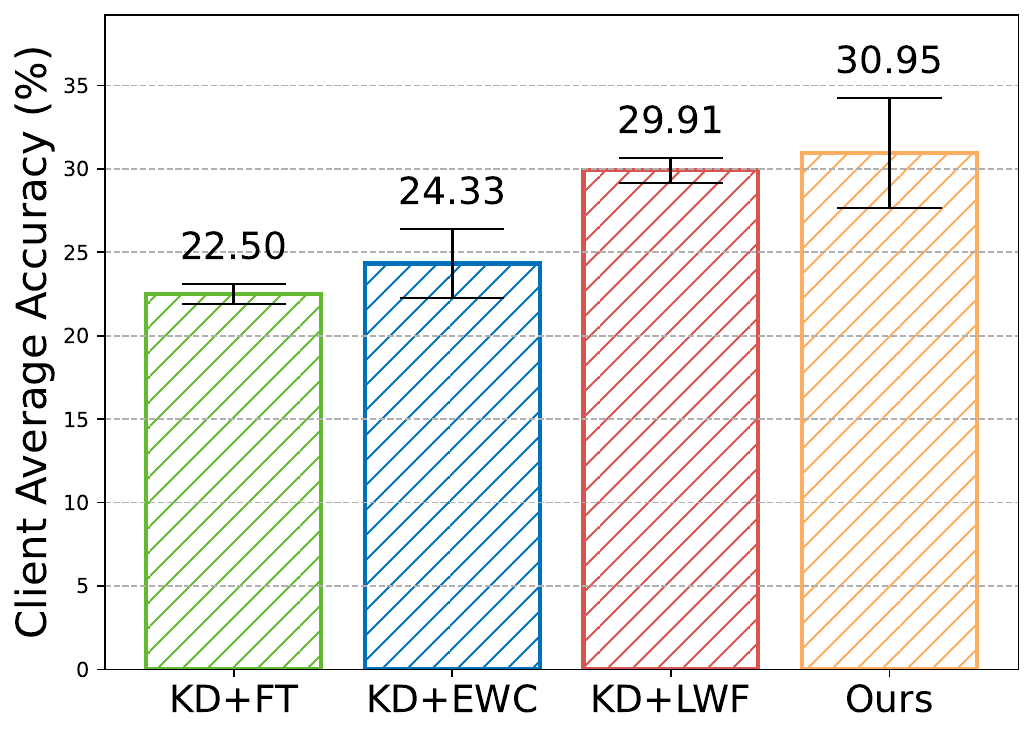} \hspace{2mm} &
            \includegraphics[width=0.43\linewidth]{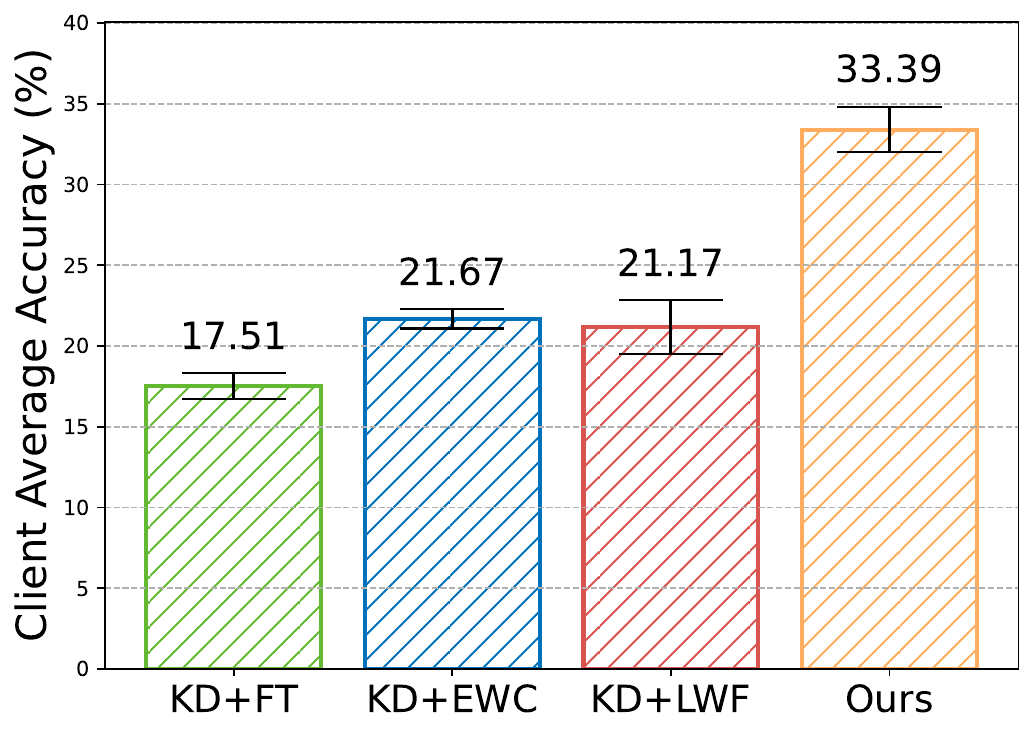}\\    
             (c) \textit{RGB} Dir($\gamma$=0.5) & (d) \textit{RGB} Dir($\gamma$=0.3)
        \end{tabular}
    \end{minipage}
    \caption{The comparison of client performance for heterogeneous model methods.}
    \label{fig:client_acc}
 \end{figure}

\subsection{Performance Evaluation}
Table.\ref{tab:overall} presents the server accuracy over all tasks seen so far, evaluated after learning each task, as well as the final forgetting measure.

The experimental data shows that our proposed \modelname \ outperforms all baselines across all settings in terms of both cumulative accuracy after learning all tasks (-\(t_3\), higher is better), and forgetting measure (F, lower is better).
Specially, on the \textit{Grayscale} datasets under \(Dir(\gamma = 0.3)\), our approach attains 80.92\% cumulative accuracy, 9.00 percentage points higher than the second-best method Target's 71.92, while reducing forgetting to 4.37, significantly outperforming all baselines. Similar gains are observed under \(Dir(\gamma = 0.5)\), where we achieve 82.99\% and 4.86, surpassing the runner-up by 6.48 percentage points  in accuracy and cutting forgetting by 6.66.  
On the more challenging \textit{RGB} datasets, our method again leads by a large margin. We obtain 74.10\% and 73.43\% cumulative accuracy, improving 23.67\% and 16.40\% compared to the closest baseline Lander, and we also achieve the lowest forgetting measure 8.40 and 6.85, respectively. This is because our approach incorporates heterogeneous models and leverages the advantages of server-side ViT fine-tuning. This highlights the effectiveness of \modelname \ in training the server model continuously on task flow.

Among the baselines, the naive approaches, FedAvg and FT+KD, achieve the poorest performance, with their high forgetting measures indicating near-complete catastrophic forgetting of previously learned knowledge. We observe that federated continual learning methods based on homogeneous models, despite using only S-type models, benefit from federated aggregation and achieve consistently better results. In particular, adaptations of classic continual learning techniques, FedEWC and FedLWF-2T, leverage regularization to significantly outperform the original FedAvg. Further improvements are observed with data-generation-based methods. Target and Lander achieve the second-best results overall; their high cumulative accuracy and low forgetting clearly show that synthetic data replay effectively preserves learned knowledge. In the baselines with heterogeneous model, the behavior of classic methods diverges. KD+EWC performs similarly to simple fine-tuning KD+FT, suggesting that parameter regularization alone is largely ineffective in this setting. On the other hand, KD+LWF successfully integrates knowledge distillation and achieves decent performance, indicating that distillation can partially bridge the heterogeneity gap. However, without dedicated optimization, its performance remains suboptimal compared to our method.

\begin{table}[t]
\caption{Ablation Study.}
\centering
\begin{tabular}{ccccccl}
\hline
\makecell[c]{\ding{172}} &
\makecell[c]{\(D_{\text{cur}}^{\text{syn}}\) \\ in \ding{173}} &
\makecell[c]{\(D_{\text{pre}}^{\text{syn}}\) \\ in \ding{173}} &
\makecell[c]{\ding{174}} &
\makecell[c]{server \\ accuracy} &
\makecell[c]{client \\ accuracy} \\ \hline
\XSolidBrush &   &   &   & -- & -- \\
 & \XSolidBrush &  &  & 47.83 & 23.17\,\(\pm\)\,2.13 \\
 & & \XSolidBrush &  & 74.30 & 21.23\,\(\pm\)\,1.32 \\
 &  &   & \XSolidBrush & -- & -- \\
\CheckmarkBold & \CheckmarkBold & \CheckmarkBold & \CheckmarkBold & 73.43 & 30.95\,\(\pm\)\,3.27 \\ \hline
\label{tab:ab}
\end{tabular}
\end{table}

Figure.~\ref{fig:performance} presents the per-task accuracy and average accuracy after learning all tasks, on different datasets and under different data heterogeneity settings. Most baselines achieve relatively high accuracy on later task but fail to retain knowledge from earlier stages, illustrating how catastrophic forgetting severely degrades overall performance. In contrast, our \modelname \ achieves the highest average accuracy and maintains high accuracy across all previously learned tasks, showcasing the effectiveness of our method in efficiently retaining old knowledge. Figure~\ref{fig:client_acc} shows the comparison of average client accuracy for methods based on heterogeneous models. \modelname\ outperforms all baselines with heterogeneous clients, owing to our approach that enhances the classification performance of clients through augmented local training and knowledge feedback from the server model.

\subsection{Ablation Study}

We conducted an ablation study on the \textit{RGB} dataset with Dir($\gamma$=0.5), as shown in Table.\ref{tab:ab}. Since module \ding{172} is the core of our paper, its ablation would affect the generation of all synthetic data; meanwhile, module \ding{174} influences the knowledge exchange of all heterogeneous models, hence it cannot be ablated. In module \ding{173}, we explore the impact of two data augmentations on our method. When the data augmentation related to the current task was removed, the model noticeably lacked the capability to counter noniid conditions, leading to a performance drop. On the other hand, when the replay of old knowledge was canceled, because the server model only acquired current-task-related knowledge from the client models without the regularization of knowledge retention, the client models could more easily learn the current knowledge and the server model could better absorb this knowledge. However, as a trade-off, the client models completely lose their ability to preseve learned knowledge, resulting in a sharp decline in performance.

\begin{table}[t]
    \centering
    \caption{Time Cost of \modelname \ on the \textit{RGB}.}
    \begin{tabular}{lccc}
    \toprule
     & Prototype Generation & Training & Data Generation \\
    \midrule
    time cost (min) & $\approx$ 5 & $\approx$ 12 & $\approx$ 3.5  \\
    rounds & 10 & 40 & - \\
    \bottomrule
    \end{tabular}
    \label{tab:timecost}
\end{table}

\begin{figure}[t]
    \centering
    \begin{minipage}[t]{0.9\linewidth}
    \centering
        \begin{tabular}{@{\extracolsep{\fill}}c@{}c@{}c@{}c@{}c@{}@{\extracolsep{\fill}}}
             \includegraphics[width=0.19\linewidth]{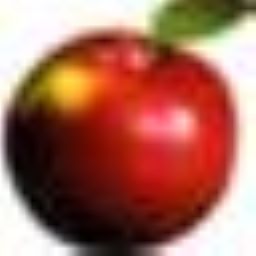} & 
            \includegraphics[width=0.19\linewidth]{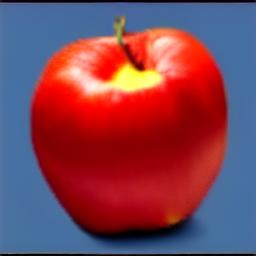} &
            \includegraphics[width=0.19\linewidth]{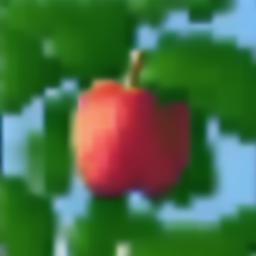}&
            \includegraphics[width=0.19\linewidth]{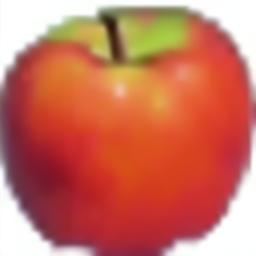}&
            \includegraphics[width=0.19\linewidth]{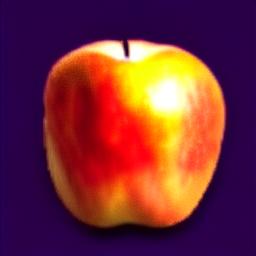}\\    
            \includegraphics[width=0.19\linewidth]{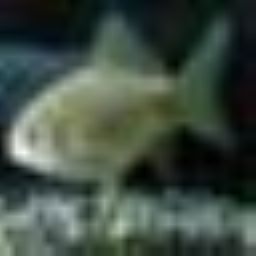} & 
            \includegraphics[width=0.19\linewidth]{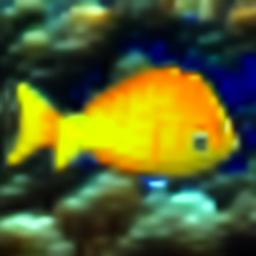} &
            \includegraphics[width=0.19\linewidth]{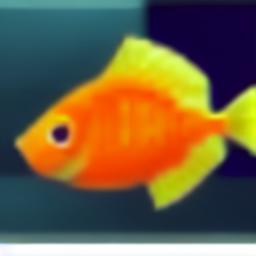}&
            \includegraphics[width=0.19\linewidth]{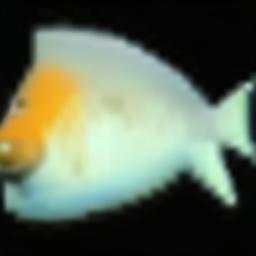}&
            \includegraphics[width=0.19\linewidth]{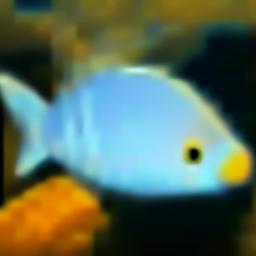}\\ 
            \includegraphics[width=0.19\linewidth]{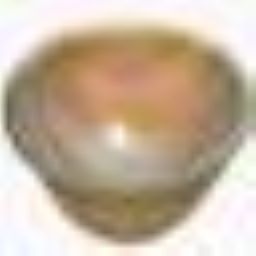} & 
            \includegraphics[width=0.19\linewidth]{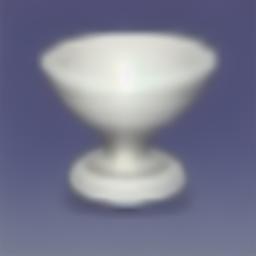} &
            \includegraphics[width=0.19\linewidth]{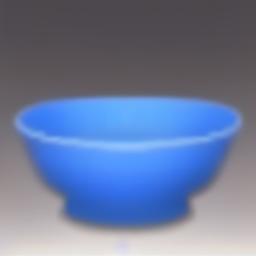}&
            \includegraphics[width=0.19\linewidth]{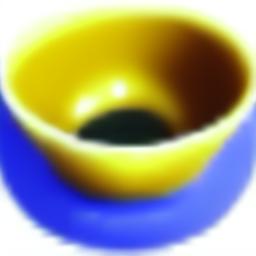}&
            \includegraphics[width=0.19\linewidth]{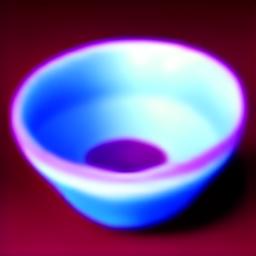}\\ 
        \end{tabular}
    \end{minipage}
    \caption{The leftmost row shows real data, and the three rows on the right show synthetic data.}
    \label{fig:syndata}
 \end{figure}

\subsection{Time Cost \& Sythetic Data Quality}
Upon encountering each new task, \modelname \ first performs federated prototype learning and then synthesizes data based on the learned prototypes, followed by model training, which includes local training and knowledge distillation. 
Notably, data generation can be performed on the server side without consuming client computational resources, and the data generated by the server-side diffusion model can be broadcast to the clients and stored for repeated use in subsequent tasks.
We report the time cost of \modelname \ on the \textit{RGB} dataset in Table.\ref{tab:timecost}. 
Specifically, the prototype training and model training duration for one client are reported, and the time spent on generating data for each class is also presented. The time spent by clients on prototype training accounts for 30\% of the total training time, which is acceptable. We show a comparison between real and generated data in Figure.\ref{fig:syndata}, where it can be seen that high-quality and highly diverse data can be generated through the lightweight prototypes.

\section{Conclusion}
In this study, we proposed \modelname, a novel federated framework that pioneers the continual collaboration of heterogeneous devices in training a cloud server model within a knowledge stream, all in a data-free manner.
Through federated prototype extraction, we leveraged pre-trained diffusion models to obtain lightweight class prototypes that capture the knowledge of the current task. These prototypes effectively facilitate data-free knowledge transfer and data-free knowledge retention.
Through data augmentation and data replay, we enhance the continuity and stability of local training, and by employing a multi-teacher distillation strategy, we effectively transfer knowledge to the server model, enabling its continual updating.
Compared with the state-of-the-arts, our algorithm improved the server accuracy by 9.00\%/6.48\%/23.67\%/16.40\% on four settings.

\bibliography{mybibliography}
\bibliographystyle{plain}

\newpage

\begin{IEEEbiography}[{\includegraphics[width=1in,height=1.25in,clip,keepaspectratio]{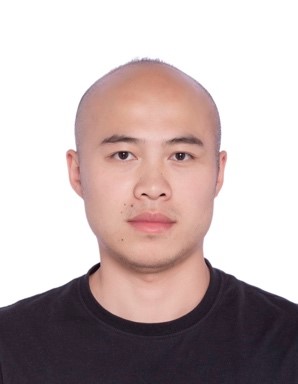}}]{Xiao Zhang} is now an associate professor in the School of Computer Science and Technology, Shandong University. His research interests include data mining, distributed  learning and federated learning. He has published more than 30 papers in the prestigious refereed journals and conference proceedings, such as IEEE Transactions on Knowledge and Data Engineering, IEEE Transactions on Mobile Computing, NeurIPS, SIGKDD, SIGIR, UBICOMP, INFOCOM, ACM MULTIMEDIA, IJCAI, and AAAI. 
\end{IEEEbiography}

\begin{IEEEbiography}[{\includegraphics[width=1in,height=1.25in,clip,keepaspectratio]{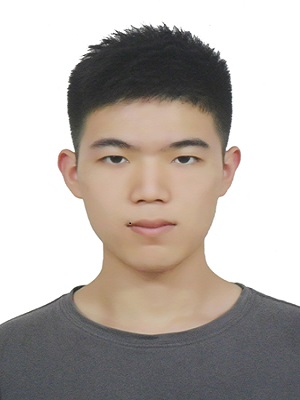}}]{Zengzhe Chen} 
 is an M.D. student in the School of Computer Science and Technology at Shandong University. He received his B.S. degree from Shandong University. His current research interests include distributed learning. 
\end{IEEEbiography}

\begin{IEEEbiography}[{\includegraphics[width=1in,height=1.25in,clip,keepaspectratio]{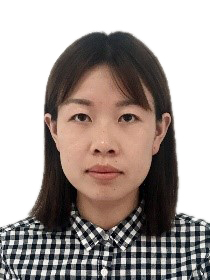}}]{Yuan Yuan} received the BSc degrees from the School of Mathematical Sciences, Shanxi University in 2016, and the Ph.D. degree from the School of  Computer Science and Technology, Shandong University, Qingdao, China, in 2021. She is currently a postdoctoral fellow at the Shandong University-Nanyang Technological University International Joint Research Institute on Artificial Intelligence, Shandong University. Her research interests include distributed computing and distributed machine learning.
\end{IEEEbiography}

\begin{IEEEbiography}[{\includegraphics[width=1in,height=1.25in,clip,keepaspectratio]{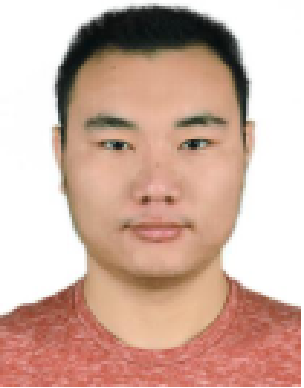}}]{Yifei Zou} received the B.E. degreefrom Computer School, Wuhan University, Wuhan,China, in 2016, and the Ph.D. degree from the Department of Computer Science, The University of Hong Kong, Hong Kong, in 2020.
He is currently an Assistant Professor with the School of Computer Science and Technology, Shandong University, Qingdao, China. His research interests include wireless networks, ad hoc networks, and distributed computing.
\end{IEEEbiography}

\begin{IEEEbiography}[{\includegraphics[width=1in,height=1.25in,clip,keepaspectratio]{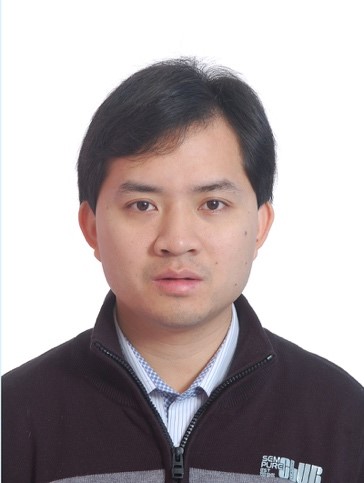}}]{Fuzhen Zhuang} received the Ph.D. degrees in the Institute of Computing Technology, Chinese Academy of Sciences, Beijing, China, in 2011. He is currently a full Professor with the Institute of Artificial Intelligence, Beihang University. He has published more than 150 papers in some prestigious refereed journals and conference proceed-ings such as Nature Comunications, the IEEE Transactions on Knowledge and Data Engineering, the IEEE Transactions on Cybernetics, the IEEE Transactions on Neural Networks and Learning Systems, the ACM Transactions on Knowledge Discovery from Data, the ACM Transactions on Intelligent Systems and Technology, Information Sciences, Neural Networks, SIGKDD, IJCAI, AAAI, TheWebConf, ACL, SIGIR, ICDE, ACM CIKM, ACM WSDM, SIAM SDM, and IEEE ICDM. His research interests include transfer learning, machine learning, data mining, multitask learning, knowledge graph and recommendation systems. He is a Senior Member of CCF. He was the recipient of the Distinguished Dissertation Award of CAAI in 2013.
\end{IEEEbiography}

\begin{IEEEbiography}[{\includegraphics[width=1in,height=1.25in,clip,keepaspectratio]{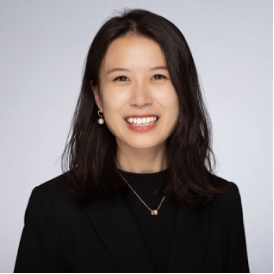}}]{Wenyu Jiao} is a Ph.D. student in Finance at the Desautels Faculty of Management, McGill University. She received her B.A. degree in Mathematics and Economics from New York University and her M.S. degree in Financial Mathematics from Johns Hopkins University. Her current research interests include asset pricing, digital assets, data automation, and the application of large language models for data analytics and decision making.
\end{IEEEbiography}

\begin{IEEEbiography}[{\includegraphics[width=1in,height=1.25in,clip,keepaspectratio]{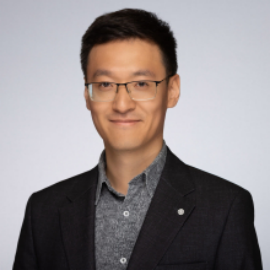}}]{Yuke Wang} is a Ph.D. student in Information System at the Desautels Faculty of Management, McGill University. He received his B.A. degree from Xi’an Jiaotong University and his Master of information studies degree from McGill University. His current research interests include the economic and societal impacts of large language models, IT innovation, and AI transformation. 
\end{IEEEbiography}

\begin{IEEEbiography}[{\includegraphics[width=1in,height=1.25in,clip,keepaspectratio]{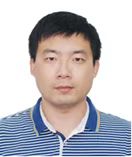}}]{Dongxiao Yu} received the B.S. degree in 2006 from the School of Mathematics, Shandong University and the Ph.D degree in 2014 from the Department of Computer Science, The University of Hong Kong. He became an associate professor in the School of Computer Science and Technology, Huazhong University of Science and Technology, in 2016. He is currently a professor in the School of Computer Science and Technology, Shandong University. His research interests include edge intelligence, distributed computing and data mining.
\end{IEEEbiography}

\vspace{11pt}

\vfill

\end{document}